# Bi-objective Optimization of Biclustering with Binary Data

**Fred Glover**
ECEE- College of Engineering and Applied Science
University of Colorado – Boulder
Boulder, CO, 80309 USA
glover@colorado.edu

**Saïd Hanafi**
LAMIH, CNRS UMR 8201
Université Polytechnique des Hauts-de-France
Valenciennes, 59313, France
Said.Hanafi@uphf.fr

**Gintaras Palubeckis**
Faculty of Informatics, Kaunas University of Technology
51368 Kaunas, Lithuania
gintaras.palubeckis@ktu.lt

**Version Febrary, 4th, 2020**

**Abstract:** Clustering consists of partitioning data objects into subsets called clusters according to some similarity criteria. This paper addresses a generalization called quasi-clustering that allows overlapping of clusters, and which we link to biclustering. Biclustering simultaneously groups the objects and features so that a specific group of objects has a special group of features. In recent years, biclustering has received a lot of attention in several practical applications. In this paper we consider a bi-objective optimization of biclustering problem with binary data. First we present an integer programing formulations for the bi-objective optimization biclustering. Next we propose a constructive heuristic based on the set intersection operation and its efficient implementation for solving a series of mono-objective problems used inside the ε-constraint method (obtained by keeping only one objective function and the other objective function is integrated into constraints). Finally, our experimental results show that using CPLEX solver as an exact algorithm for finding an optimal solution drastically increases the computational cost for large instances, while our proposed heuristic provides very good results and significantly reduces the computational expense.

**Keywords:** Bi-clustering, Bi-objective Optimization, Biclique, ε-Constraint Method.

# 1. Introduction

Clustering is a technique that involves the grouping a set of objects in such a way that objects in the same cluster are more similar (in some sense) to each other than to those in other clusters, exploiting the structure of data without requiring the assumptions common to most statistical approaches. Called unsupervised learning or unsupervised classification, clustering is used for data analysis in many fields, including data mining, machine learning, pattern recognition, image analysis and bioinformatics. Such techniques have been designed for a variety of data types — homogeneous and nonhomogeneous numerical data, categorical data, binary data. This paper addresses a generalization called *quasi-clustering* that allows overlapping of clusters, and which we link to biclustering

Formally, let an input data set of $r$ objects (samples) and $n$ features (attributes) be given as a rectangular matrix $A = (a_{ij})_{r \times n}$, where the value $a_{ij}$ is the expression of the $i^{th}$ object in the $j^{th}$ feature. Clustering consists of partitioning the data objects into disjoint subsets (clusters) according to some similarity criteria. Biclustering simultaneously groups the objects and features so that a specific group of objects has a special group of features. More precisely, a biclustering technique identifies a subset of objects (rows) that exhibit similar patterns on a subset of features (columns) in an input data matrix $A$.

There exist different names used to designate the biclustering domain, including co-clustering, bidimensional clustering, two-mode clustering, bimodal cluster analysis, coupled two-way clustering, two-way clustering and direct clustering, among others. The term biclustering was introduced by Mirkin (1996) and was notably applied by Cheng and Church (2000) to analyze gene expression data, which significantly contributed to the popularization of biclustering techniques. However, the biclustering model can be traced at least as far back as the works of Malgrange (1962) and was treated in the work of Hartigan (1972).

A biclustering algorithm looks for a set of biclusters such that each bicluster satisfies certain characteristics of homogeneity, giving rise to versions such as : i) biclusters with constant values, ii) biclusters with constant values in columns or rows, iii) biclusters with coherent values, and iv) biclusters with coherent evolutions. (Definitions and examples of these various bicluster types can be found in Madeira and Oliveira, 2004). Most of the literature on biclusters presents heuristic or exact algorithms for enumerating all maximal biclusters. In the general case, the biclustering problem is NP-hard (Cheng and Church 2000).

Clustering is closely connected to combinatorial optimization and graph theory. In particular, biclustering is related to bipartite graph partitioning (Dhillon (2001), Zha et al. (2001), Zhao et al. (2008), Fan et al. (2010)). An interesting connection between data matrices and graph theory can be established as follows. The data matrix $A = (a_{ij})$ can be viewed as a weighted bipartite graph $G = (R, N, E)$ where each node $i \in R$ corresponds to a row and each node $j \in L$ corresponds to a column where an edge $(i,j) \in E$ has weight $a_{ij}$. A biclique of bipartite $G$ is a subgraph of $G$ that is also a complete bipartite graph. A biclique of a bipartite graph therefore corresponds to a bicluster of the associated matrix. Finding a biclique of a maximum number of vertices can be done in polynomial time (Garey, M. R., & Johnson, D. S. (1979)), while finding a biclique of a maximum number of edges is NP-complete (Peeters (2003)). There exist several computationally challenging problems related to bicliques such as enumerating maximal biclique subgraphs and covering the edges of a bipartite graph by bicliques. Other important applications of those problems arise in the context of data compression (Agarwal et al. 1994), automata and language theories (Froidure (1995)), graphs (Fishburn, and Hammer, (1996), partial orders (Habib et al. (1997)).

In recent years, biclustering has proved to be a powerful data analysis technique due to its wide success in various application domains, particularly in microarray and gene expression analysis, computational biology, biomedicine, text mining, pattern discovery, tokens and contexts in natural language processing, data exploration, marketing, web search, collaborative filtering and many other applications. Useful reviews on biclustering techniques and their applications can be found in Cheng and Church (2000), Madeira and Oliveira (2004), Tanay et al. (2005), Busygin et al. (2008), Fan et al. (2010). The reader is also referred to a recent survey by Pavlopoulos et al. (2018), that additionally covers a broad spectrum of problems on bipartite graphs. In the recent past, several authors have investigated a special case of the bipartite graph partitioning problem where the data matrix $A$ is 0-1 valued and thus represented by an unweighted graph, notably in Prelić et al. (2006), Dolnicar et al. (2012), Orzechowski and Boryczko (2012), and Wang et al. (2016). Analysis of the literature shows that various variants of the biclustering problem typically are treated as single-objective problems. Considering this observation, our motivation is to study a bi-objective version of the problem where the goal is to maximize the size of each side of the biclique.

Let $x = (x_1, \ldots, x_n)$ be a vector of binary variables of dimension $n$, i.e. $x_j \in \{0,1\}, j \in N = \{1, \ldots, n\}$, and let $R$ denote a reference set of $x$ vectors (solutions) which we represent by $R = \{x(1), \ldots, x(r)\}$. For $S \subset R$, define the *intersect sets* $N_0(S)$, $N_1(S)$ and $N_{01}(S)$ by

$$N_0(S) = \{j \in N : x_j = 0 \ for\ every\ x \in S\}$$

$$N_1(S) = \{j \in N : x_j = 1 \ for\ every\ x \in S\}$$

$$N_{01}(S) = N_0(S) \cup N_1(S)$$

We are interested in identifying sets $S \subset R$ that maximize the size of these intersect sets, i.e., that maximize $|N_0(S)|$ or $|N_1(S)|$ or $|N_{01}(S)|$. More precisely, the goal of identifying sets $S \subset R$ that maximize simultaneously $|S|$ the size of the set $S$ and $|N_v(S)|$ the size of intersect set can be expressed as that of solving the following bi-objective optimization problem

$$(G_v) \quad \begin{cases} max & g_v(S) = (|S|, |N_v(S)|) \\ s.t. & S \subseteq R \end{cases}$$

We will call this formulation the *Set formulation*.

The remainder of this paper is organized as follows. In Section 2, we provide basic definitions for bi-objective optimization problems and we present an ε-constraint Method for a bi-objective optimization problem which solves a series of mono-objective problems by keeping only one objective function and the other objective function is integrated into constraints. Section 3 presents an integer programing and biclique formulations for the bi-objective optimization biclustering problem. Section 4 is dedicated to approximate algorithms for solving the mono-objective problems, proposing an effective constructive heuristic and its efficient implementation. Section 5 illustrates the running of the proposed algorithms on a small instance. Advanced considerations are covered in Section 6, including a generalization called quasi-clustering that allows overlapping of clusters. Section 7 presents the generation of the instances and the results of extensive computational experiments. Concluding remarks are given in Section 8.

## 2. ε-Constraint Method

First, we recall some basic definitions for bi-objective optimization problems. The objective space is defined by $Y_v = \{(g_v^1(S), g_v^2(S)) : S \subseteq R\}$. Since in general, there is no feasible solution which minimizes the two objectives $g_v^1(S)$ and $g_v^2(S)$ simultaneously, we search for an acceptable trade-off between them. This compromise is defined by a dominance relation which corresponds to a partial order on the objective space $Y_v$.

**Definition 1.** (*Pareto dominance*). Let $g_v$ and $g_v'$ be two solutions in the objective space $Y_v$ of a bi-objective problem. We say that $g_v$ dominates $g_v'$, denoted by $g_v \succ g_v'$, if and only if $g_v^k \geq g_v'^k$, for $k = 1, 2$, with at least one inequality being strict.

We remark that the widely-used concept of Pareto optimization, which appeared at the end of the 19th century (Pareto, 1896), is useful for finding the compromise solutions in our present context. We apply the common terminology of referring to the set of all non-dominated solutions in the space of objectives as the optimal Pareto front (Allais, 1968).

**Definition 2.** (*Pareto efficiency*). A solution $S \subseteq R$ is called Pareto efficient, if and only if no solution $S' \subseteq R$ exists such that $g_v(S) \succ g_v(S')$. The efficient set is denoted by $E_v^* = \{S \subseteq R : S \text{ is Pareto efficient}\}$ and the Pareto front is denoted by $F_v^* = \{g_v(S) : S \in E_v^*\}$.

The efficient set $E_v^*$ and Pareto front $F_v^*$ contain all the Pareto efficient solutions and all the non-dominated points in the objective space, respectively. In other words, $S^* \subseteq R$ is efficient if there is no other feasible solution $S \subseteq R$ which leads to an improvement in some criterion without simultaneous deterioration in at least one other.

The Ideal and Nadir points are upper and lower bounds on non-dominated points. These points give an indication of the range of the values which non-dominated points can attain.

**Definition 3.** (*Ideal and Nadir points*). The point $g_v^I = (g_v^{I,1}, g_v^{I,2})$ with $g_v^{I,k} = \max\{g_v^k(S) : S \subseteq R\}$ for $k = 1, 2$, is called the Ideal point. The point $g_v^N = (g_v^{N,1}, g_v^{N,2})$ with $g_v^{N,k} = \min\{g_v^k(S) : S \in E_v^*\}$ for $k = 1, 2$, is called Nadir point.

For the bi-objective problem $G_v$, the cordinates of Ideal and Nadir points can be determined as follows:

$$g_v^{I,1} = \max\{|S| : S \subseteq R\} = |R| = r$$

$$g_v^{I,2} = \max\{|N_v(S)| : S \subseteq R\} = \begin{cases} \max\{|N_v(x)| : x \in R\} & if\ v \in \{0,1\} \\ n & if\ v = 01 \end{cases}$$

$$g_v^{N,1} = \min\{|S| : |N_v(S)| = g_v^{I,2}, S \subseteq R\} = 1$$

$$g_v^{N,2} = \min\{|N_v(S)| : |S| = r, S \subseteq R\} = |N_v(R)|.$$

The determination of the Ideal and Nadir points yield the following consequences:
   i) The points $(g_v^{I,1}, g_v^{N,2})$ and $(g_v^{N,1}, g_v^{I,2})$ are efficient points, i.e.
   $\{(g_v^{I,1}, g_v^{N,2}), (g_v^{N,1}, g_v^{I,2})\} \subseteq F^*$.
   ii) The evaluation vector $(g_v^1(S), g_v^2(S))$ of a solution $S \in E_v^*$ is bounded as follows:
   $$g_v^{N,1} \leq g_v^1(S) \leq g_v^{I,1}, and\ g_v^{N,2} \leq g_v^2(S) \leq g_v^{I,2}$$
Hence, for the bi-objective problem $G_v$, we have:
$$\{(r, |N_v(R)|), (1, g_v^{I,2})\} \subseteq F_v^*$$
$\forall S \in E_v^*, we\ have\ 1 \leq g_v^1(S) \leq r, and\ |N_v(R)| \leq g_v^2(S) \leq g_v^{I,2}$

The ε-constraint method, introduced by Haimes, Lasdon and Wismer (1971), chooses a single - objective to be optimized while the other objective is treated as a constraint. More specifically, the bi-objective optimization problem $G_v$ is replaced by one of the two parametric problems $G_v^1$ and $G_v^2$, i.e.

$$(G_v^1(L_v^1)) \quad \begin{cases} max & g_v^1(S) = |S| \\ s.t. & g_v^2(S) = |N_v(S)| \geq L_v^1 \\ & S \subseteq R \end{cases}$$

$$(G_v^2(L_v^2)) \quad \begin{cases} max & g_v^2(S) = |N_v(S)| \\ s.t. & g_v^1(S) = |S| \geq L_v^2 \\ & S \subseteq R \end{cases}$$

The ε-constraint method is justified by the following properties:

- An optimal solution $S_v^*$ of $G_v^k(L_v^k)$ for some $k$, is weakly efficient, i.e. $\nexists S \subseteq R: g_v^k(S) \geq g_v^k(S_v^*)$ for $k = 1,2$.
- If there exists an unique optimal solution $S_v^*$ of $G_v^k(L_v^k)$ for some k, then $S_v^*$ is strictly efficient i.e. $\nexists S \subseteq R: g_v^k(S) > g_v^k(S_v^*)$ for $k = 1,2$, (hence it is efficient).
- A set $S \subseteq R$ is efficient if and only if there exists $(L_v^1, L_v^2)$ such that $S$ is an optimal solution of $G_v^k(L_v^k)$ for all $k = 1, 2$.

The extreme point that is computed is then used to determine the bound on the objectives, and this is repeated until there are no new solutions left.

The ε-constraint method is easy to implement, but it requires potentially high computational cost (many runs may be required).

### ε-Constraint Algorithm for $G_v$ Problem

Choose $p \in \{1,2\}$;

Compute the Ideal point $g_v^I = (g_v^{I,1}, g_v^{I,2})$ and Nadir point $g_v^N = (g_v^{N,1}, g_v^{N,2})$

Set $F = \{(g_v^{I,p}, g_v^{N,3-p})\}$ and $L_v^{3-p} = g_v^{N,3-p} + \varepsilon, (\varepsilon = 1)$

**While $L_v^{3-p} \leq g_v^{I,3-p}$ do**

Solve $G_v^p(L_v^{3-p})$ to obtain an optimal solution $S^*$ with the values $(g_v^{*,p}, g_v^{*,3-p})$

Set $F = F \cup \{(g_v^{*,p}, g_v^{*,3-p})\}$ and $L_v^{3-p} = g_v^{*,3-p} + \varepsilon$

**Endwhile**

Remove dominated points from $F$ if required.

**Return** $F$.

## 3. Alternative Formulations of Problems $G_v$, $G_v^1$ and $G_v^2$

### 3.1. Integer programing Formulation

For a given 0-1 vector $x$, the solution $\bar{x}$ denotes the complemented vector of $x$ given by $\bar{x}_j = 1 - x_j, j \in N$. For each subset $J$ of $N$, $x_J$ denotes the subvector of $x$ defined by $x_J = (x_j)_{j \in J}$. A vector of ones of arbitrary dimension will be denoted by $e$. Hence, we have $\bar{x} = e - x$.

The intersect sets $N_0(S)$, $N_1(S)$ and $N_{01}(S)$, can be redefined by

$$N_0(S) = \{j \in N: ex_S = \sum_{x \in S} x_j = 0\},$$

$$N_1(S) = \{j \in N: e\bar{x}_S = \sum_{x \in S} \bar{x}_j = 0\} = \{j \in N: ex_S = |S|\},$$

$$N_{01}(S) = \{j \in N: ex_S \times e\bar{x}_S = 0\}.$$

To formulate the problems $G_v^1$ and $G_v^2$ as Integer Programs, we introduce the binary variables $y_i$ for $i \in R$ and $z_j^v$ for $j \in N$, $v \in \{0,1\}$ with the following meaning

$$y_i = \begin{cases} 1 & if\ i \in S, i.e.\ x(i)\ is\ selected\ from\ R \\ 0 & Otherwise \end{cases},$$

$$z_j^v = \begin{cases} 0 & if\ j \in N_v(S) \\ 1 & Otherwise \end{cases}.$$

With these binary variables, the cardinalities of a subset $S \subseteq R$ and its intersect set $N_v(S)$ can be expressed as

$$|S| = \sum_{i \in R} y_i,$$

$$|N_v(S)| = \sum_{j \in N}(1 - z_j^v).$$

Hence, for $v \in \{0,1\}$ the constraint (1) can be expressed as

$$\sum_{j \in N}(1 - z_j^v) \geq L_v^1 \qquad (CL1\text{-}v)$$

and for $v = 01$, the constraint (1) becomes

$$\sum_{j \in N}(2 - z_j^0 - z_j^1) \geq L_{01}^1 \qquad (CL1\text{-}01)$$

Each component $j \in N$ is a member of the set $N_0(S)$, if and only if $\sum_{i \in R} x(i)_j y_i = 0$, hence $z_j^0 = 0$. This can be expressed by the following constraints linking the variables $y_i$ and $z_j^0$:

$$z_j^0 \leq \sum_{i \in R} x(i)_j y_i \leq r z_j^0 \qquad (CS\text{-}0)$$

Similarly, each component $j \in N$ is member the set $N_1(S)$, if and only if $\sum_{i \in R}(1 - x(i)_j) y_i = 0$, hence $z_j^1 = 0$. This can be expressed by:

$$z_j^1 \leq \sum_{i \in R}(1-x(i)_j)y_i \leq rz_j^1 \qquad (CS\text{-}1)$$

The IP formulations for the problems $G_v^1$ for $v \in \{0,1\}$ can be stated as follows

$$(IP_v^1) \quad \begin{cases} max & y_0 = \sum_{i \in R} y_i \\ s.t. & (CL1-v), (CS-v) \\ & y \in \{0,1\}^r, z^0 \in \{0,1\}^n \end{cases}$$

and for $v = 01$ the IP can be expressed as

$$(IP_{01}^1) \quad \begin{cases} max & y_0 = \sum_{i \in R} y_i \\ s.t. & (CL1-01), (CS-0), (CS-1) \\ & y \in \{0,1\}^r, z^0, z^1 \in \{0,1\}^n \end{cases}$$

For the problem $G_v^2$, the constraint (2) can be expressed as

$$\sum_{i \in R} y_i \geq L_v^2 \qquad (CL2\text{-}v)$$

Hence, the IP formulations for the problems $G_v^2$ for $v \in \{0,1\}$ can be stated as follows

$$(IP_v^2) \quad \begin{cases} max & y_0 = \sum_{j \in N}(1-z_j^v) \\ s.t. & (CL2-v), (CS-v) \\ & y \in \{0,1\}^r, z^0 \in \{0,1\}^n \end{cases}$$

and for $v = 01$ the IP can be expressed as

$$(IP_{01}^2) \quad \begin{cases} max & y_0 = \sum_{j \in N}(1-z_j^0) + (1-z_j^1) \\ s.t. & (CL2-01), (CS-0), (CS-1) \\ & y \in \{0,1\}^r, z^0, z^1 \in \{0,1\}^n \end{cases}$$

Let $A$ be the $r \times n$ matrix with components $x(i)_j$ for $i \in R$ and $j \in N$ and $\bar{A}$ be the $r \times n$ matrix with components $\bar{x}(i)_j = 1 - x(i)_j$ for $i \in R$ and $j \in N$. Using those matrices, the integer program $IP^v$ formulation for problems $G_v$ can be expressed as

for $v \in \{0,1\}$

$$(IP^v) \quad \begin{cases} max & (y_0, z_0) = (ey, e\bar{z}^v) \\ s.t. & z^v \leq (A + v(\bar{A} - A))y \leq rz^v \\ & y \in \{0,1\}^r, z^0 \in \{0,1\}^n \end{cases}$$

And for $v = 01$

$$(MIP^{01}) \quad \begin{cases} max & (y_0, z_0) = (ey, e(\bar{z}^0 + \bar{z}^1)) \\ s.t. & z^k \leq (A + k(\bar{A} - A))y \leq rz^k \quad k \in \{0,1\} \\ & y \in \{0,1\}^r, z^k \in \{0,1\}^n \quad k \in \{0,1\} \end{cases}$$

**Remark 1**: Let $f_j = \sum_{i \in R} x(i)_j$ and $\bar{f}_j = r - f_j$, hence the constraints (CS-0) and (CS-1) can be strengthened as follows

$$z_j^0 \leq \sum_{i \in R} x(i)_j y_i \leq f_j z_j^0 \quad \text{(CS'-0)}$$

$$z_j^1 \leq \sum_{i \in R} (1 - x(i)_j) y_i \leq \bar{f}_j z_j^1 \quad \text{(CS'-1)}$$

### 3.2. Bicliques Formulation

In this section, we redefine problems $G_v$, $G_v^1$ and $G_v^2$ using graph terminology. First, we begin by defining the bipartite graph $G^v = (R, N, E^v)$, for $v \in \{0,1\}$, where $E^v = \{(i,j): i \in R, j \in N$ such that $x(i)_j = v\}$. Note that the graph $G^0$ is the bipartite graph complement of the graph $G^1$ and vice-versa. A *biclique* $B^v(S, M)$ of $G^v$ is a complete bipartite subgraph of $G^v$ induced by $S \subseteq R$ and $M \subseteq N$ such that $S$ and $M$ are nonempty. Trivially, a biclique $B^v(S, M)$ of $G^v$ has $|S| + |M|$ vertices and $|S| \times |M|$ edges. Hence, our problem $G_v$ for $v \in \{0,1\}$, can be expressed as follows

$$(G_v) \quad \begin{cases} \max & f_v(S, M) = (|S|, |M|) \\ \text{s.t.} & B^v(S, M) \text{ is a biclique of } G^v \end{cases}$$

The situation is slightly different for $v = 01$. In this case, feasible solutions of problem $G_{01}$ are the union of the bicliques $B^v(S, M)$ of $G^v$ for $v \in \{0,1\}$. Hence, the problem $G_{01}$ can be expressed as follows

$$(G_{01}) \quad \begin{cases} \max & f_{01} = (|S^0 \cup S^1|, |M^0 \cup M^1|) \\ \text{s.t.} & B^0(S^0, M^0) \text{ is a biclique of } G^0 \\ & B^1(S^1, M^1) \text{ is a biclique of } G^1 \end{cases}$$

A biclique $B^v(S, M)$ is a *maximal biclique* of $G^v$ if there exists no biclique $B^v(S', M')$ such that, $S \subseteq S'$, $M \subseteq M'$ and $(S', M') \neq (S, M)$. Given a maximal biclique $B^v(S, M)$, it is clear that $M = N_v(S)$, hence $|N_v(S)| = |M|$.

**Remark 3**: Each efficient solution of the problem $G_v$ corresponds to a maximal biclique of a bipartite graph $G^v$, although a maximal biclique may not correspond to an efficient solution.

## 4. Approximate Algorithms for Problems $G_v^1$ and $G_v^2$

We approach these problems by a constructive search process which has the same general form for each. It is useful to organize the search for an $S$ that solves these problems by selecting some solution $x^1 = x(h) \in R$ as a "first solution" for $S$. Such an approach limits the possible remaining solutions that can belong to $S$ and aids the search for an appropriate $S$. The strategy of beginning with a first solution $x^1$ is also motivated by the fact that every $S$ contains at least one $x \in R$, and several different solutions $x \in R$ may reasonably give rise to the same $S$. Therefore, we may usefully discover different sets $S$ by selecting a new $x^1$ that lies outside sets previously generated to explore farther. This type of strategy is additionally appealing because the sets $S$ generated will typically have different properties and provide a boost to diversification.

Our constructive algorithm successively enlarges a set $S$ (which begins as $S = \{x^1\}$ for a selected solution $x^1 \in R$) by adding a single solution $x$ to create a new set $S$. The process relies on the following relationship, which is implied by our definitions:

$$N_v(S) = \cap(N_v(x): x \in S)$$

This identity motivates the terminology that calls $N_v(S)$ an intersect set.

We restate Problems $G_v^1$ and $G_v^2$ by including the stipulation that $S$ contains a given solution $x(h)$ in $R$ and using the alternative representation of Section 3. As before, let $L_v^1$ denote a chosen lower limit on the size of $N_v(S)$, and let and $L_v^2$ denote a chosen lower limit on the size of $S$, for $v = 0$ or 1 or 01. Then we state:

**Problem $G_v^1(x(h))$**: For a specified $x(h) \in R$, solve

$$(G_v^1(x(h))) \quad \begin{cases} max & |S| \\ s.t. & |N_v(S)| \geq L_v^1 \\ & x(h) \in S \\ & S \subseteq R \end{cases}$$

**Problem $G_v^2(x(h))$**: For a specified $x(h) \in R$, solve

$$(G_v^2(x(h))) \quad \begin{cases} max & |N_v(S)| \\ s.t. & |S| \geq L_v^2 \\ & x(h) \in S \\ & S \subseteq R \end{cases}$$

Let $ov(P)$ denotes the optimal value of an optimization problem P. It is obvious that we have

$$ov(G_v^1) = \max\left\{ov\left(G_v^1(x(h))\right): x(h) \in R\right\},$$

$$ov(G_v^2) = \max\left\{ov\left(G_v^2(x(h))\right): x(h) \in R\right\}.$$

### 4.1. Integer Programming formulations for Problems $G_v^1(x(h))$ and $G_v^2(x(h))$

Straightforward IP formulations of the problems $G_v^1(x(h))$ and $G_v^2(x(h))$ can be derived directly from $IP_v^1$ and $IP_v^2$ respectively by setting $y_h = 1$. However, we can also exploit the fact that we know the value of $x(h)_j$ for $j \in N$. Hence, the IP formulations for the problems $G_v^1(x(h))$ for $v \in \{0,1\}$ can be stated as follows:

$$(IP_v^1(x(h))) \quad \begin{cases} max & y_0 = \sum_{i \in R-h} y_i \\ s.t. & \sum_{j \in N_v(x(h))} (1 - z_j^v) \geq L_v^1 \\ & z_j^v \leq \sum_{i \in R-h} (x(i)_j + v(1 - 2x(i)_j)y_i) \leq rz_j^v \quad j \in N_v(x(h)) \\ & y \in \{0,1\}^{r-1}, z^v \in \{0,1\}^{|N_v(x(h))|} \end{cases}$$

and for $v = 01$ the $IP_{01}^1$ formulation becomes

$$(IP_{01}^1(x(h))) \begin{cases} \max & y_0 = \sum_{i \in R-h} y_i \\ s.t. & \sum_{j \in N} (2 - z_j^0 - z_j^1) \geq L_{01}^1 \\ & z_j^0 \leq \sum_{i \in R-h} x(i)_j y_i \leq r z_j^0 \quad j \in N_0(x(h)) \\ & z_j^1 \leq \sum_{i \in R-h} (1 - x(i)_j) y_i \leq r z_j^1 \quad j \in N_1(x(h)) \\ & y \in \{0,1\}^{r-1}, z^0, z^1 \in \{0,1\}^{|N_v(x(h))|} \end{cases}$$

The corresponding IP formulations for the problems $G_v^2(x(h))$ for $v \in \{0,1\}$ can be stated as:

$$(IP_v^2(x(h))) \begin{cases} \max & y_0 = \sum_{j \in N_v(x(h))} (1 - z_j^v) \\ s.t. & \sum_{i \in R-h} y_i \geq L_v^2 - 1 \\ & z_j^v \leq \sum_{i \in R-h} (x(i)_j + v(1 - 2x(i)_j) y_i) \leq r z_j^v \quad j \in N_v(x(h)) \\ & y \in \{0,1\}^{r-1}, z^v \in \{0,1\}^{|N_v(x(h))|} \end{cases}$$

and for $v = 01$ the $IP_{01}^2$ formulation becomes

$$(IP_{01}^2(x(h))) \begin{cases} \max & y_0 = \sum_{j \in N} (2 - z_j^0 - z_j^1) \\ s.t. & \sum_{i \in R-h} y_i \geq L_{01}^2 - 1 \\ & z_j^0 \leq \sum_{i \in R-h} x(i)_j y_i \leq r z_j^0 \quad j \in N_0(x(h)) \\ & z_j^1 \leq \sum_{i \in R-h} (1 - x(i)_j) y_i \leq r z_j^1 \quad j \in N_1(x(h)) \\ & y \in \{0,1\}^{r-1}, z^0, z^1 \in \{0,1\}^{|N_v(x(h))|} \end{cases}$$

**Remark 2**: We identify several properties that can be useful for preprocessing or that can provide valid inequalities for the proposed integer programs $IP^v$, $IP_v^1$ and $IP_v^2$:

- For $v \in \{0, 01\}$, if $f_j = 0$ then $j \in N_v(S)$ for all $S \subseteq R$. Hence the variable $z_j^0$ can be fixed to 0 in the $IP^v, IP_v^1$ and $IP_v^2$.
- For $v \in \{1, 01\}$, if $f_j = r$, i.e. $\bar{f}_j = 0$, implies $j \in N_v(S)$ for all $S \subseteq R$. Hence the variable $z_j^1$ can be fixed to 0 in the $IP^v, IP_v^1$ and $IP_v^2$.

- For $v = 0$, each variable $z_j^0$ can be fixed to 1 for $j \in N_1(x(h))$ in the $IP_0^1(x(h))$.
- For $v = 1$, each variable $z_j^1$ can be fixed to 1 for $j \in N_0(x(h))$ in the $IP_1^1(x(h))$.
- For $v \in \{0, 01\}$, let $S$ be a feasible solution for $IP_0^1(x(h))$, (i.e. $|N_v(S)| \geq L_v^1$), if $j \in N_0(x(h))$ and $\bar{f}_j < |S|$ then the variable $z_j^0$ can be fixed to 1.
- For $v \in \{1, 01\}$, let $S$ be a feasible solution for $IP_1^1(x(h))$, (i.e. $|N_v(S)| \geq L_v^1$), if $j \in N_1(x(h))$ and $f_j < |S|$ then the variable $z_j^0$ can be fixed to 1.
- In $IP_v^1(x(h))$, the binary variable $y_i$ can be fixed to 0 if
$$\begin{cases} |N_v(x(h)) \cap N^=(x(h), x(i))| < L_v^1 & for\ v \in \{0, 1\} \\ |N^=(x(h), x(i))| < L_v^1 & for\ v = 01 \end{cases}$$
where $N^=(x(h), x(i)) = \{j \in N: x(h)_j = x(i)_j\}$
- For $v \in \{0, 1\}, j \in N_v(S)$ implies $|S| \leq, \bar{f}_j + v(r - 2\bar{f}_j)$.
- For $v = 01, j \in N_v(S)$ implies $|S| \leq, \max(f_j, \bar{f}_j)$.

## 4.2. Heuristics for Problems $G_v^1(x(h))$ and $G_v^2(x(h))$

The heuristics described in this section for problems $G_v^1(x(h))$ and $G_v^2(x(h))$ provide an additional option of being used in conjunction with the ε-Constraint Algorithm for problem $G_v$. The 'while' loop in the ε-Constraint Algorithm applies an exact algorithm for finding an optimal solution which can drastically increase the computational cost for large $n$. Our proposed constructive techniques can be used to replace this exact algorithm to significantly reduce the computational expense.

### 4.2.1. An alternative representation

To visualize the constructive process subsequently described, it is convenient to define an operation $\cap_o$ on vectors $x \in S$ that gives outcomes equivalent to the set intersection operation $\cap$ on the sets $N_v(x)$. The components $x_j$ of the $x$ vectors that are operated on by $\cap_o$ can take a value # in addition to the values 0 and 1, where $x_j = $ # indicates that $x_j$ is irrelevant to defining the intersection.

The vector $x = x' \cap_o x''$ generated from two vectors $x'$ and $x''$ can be identified by reference to a corresponding operation, designated by the same symbol $\cap_o$, which is carried out on each component of $x'$ and $x''$; that is

$$x_j = x_j' \cap_o x_j'', j \in N.$$

This operation is made precise by the following rules.

$$x_j = \begin{cases} 0 & if\ x_j' = x_j'' = 0 \\ 1 & if\ x_j' = x_j'' = 1 \\ \# & otherwise, \text{hence if } x_j' \neq x_j'' \text{ or if } x_j' = \# \text{ or } x_j'' = \# \end{cases}$$

This operation be also coded simply as follows

$$x_j = \begin{cases} x_j' & if\ x_j' = x_j'' \\ \# & otherwise \end{cases}$$

Note that if the value # is coded by a fractional real number $\alpha \in ]0,1[$, then $x_j$ can be expressed simply as:

$$x_j = \frac{x_j' + x_j''}{2}.$$

By implication $\cap_o$ is commutative and associative, i.e., for values $t, u, w \in \{0, 1, \#\}$

$$t \cap_o u = u \cap_o t \text{ and } t \cap_o (u \cap_o w) = (t \cap_o u) \cap_o w.$$

These same relationships hold when $t$, $u$ and $w$ are vectors, using the vector form of $\cap_o$.

By these definitions we can write

$$\cap_o(S) = \cap_o(x: x \in S).$$

By convention, when $S$ consists of a single vector $x$, we define $\cap_o(S) = \cap_o(x) = x$.

**Construction Example.** Consider the case where $S$ consists of the following four vectors:

| j | 1 | 2 | 3 | 4 | 5 | 6 | 7 | 8 | 9 | 10 | 11 |
|---|---|---|---|---|---|---|---|---|---|----|----|
| $x^1$ | 0 | 1 | 0 | 0 | 1 | 1 | 1 | 0 | 1 | 0 | 0 |
| $x^2$ | 1 | 1 | 0 | 0 | 1 | 1 | 1 | 1 | 0 | 0 | 0 |
| $x^3$ | 0 | 1 | 0 | 1 | 1 | 1 | 0 | 1 | 1 | 0 | 0 |
| $x^4$ | 1 | 1 | 0 | 1 | 0 | 1 | 0 | 1 | 1 | 1 | 0 |

Then we can generate a vector $z = \cap_o(S)$ in four successive stages to identify vectors $z^1$, $z^2$, $z^3$ and $z^4$, as shown below, where the final vector $z^4$ is the vector $z = \cap_o(S)$. In sequence, these stages yield $z^1 = x^1$, $z^2 = z^1 \cap_o x^2$, $z^3 = z^2 \cap_o x^3$) and finally $z^4 = z^3 \cap_o x^4$.

| j | 1 | 2 | 3 | 4 | 5 | 6 | 7 | 8 | 9 | 10 | 11 |
|---|---|---|---|---|---|---|---|---|---|----|----|
| $z^1$ | 0 | 1 | 0 | 0 | 1 | 1 | 1 | 0 | 1 | 0 | 0 |
| $z^2$ | # | 1 | 0 | 0 | 1 | 1 | 1 | # | # | 0 | 0 |
| $z^3$ | # | 1 | 0 | # | 1 | 1 | # | # | # | 0 | 0 |
| $z^4$ | # | 1 | 0 | # | # | 1 | # | # | # | # | 0 |

We observe that once $z_j$ receives the value # in one of these vectors $z^k$, it continues to receive this value in all remaining vectors for larger values of $k$.

We now complete the connection with the intersect set $N_v(S)$ defined earlier. We call $z = \cap_o(S)$ the *vector analog* of $S$ and use it as an easy way to identify $N_v(S)$ by reference to the values $z_j$ taken by the components of $z$.

$$N_0(S) = \{j \in N: z_j = 0\}$$

$$N_1(S) = \{j \in N: z_j = 1\}$$

$$N_{01}(S) = \{j \in N: z_j = 0 \text{ or } z_j = 1\}$$

$$= \{j \in N: z_j \neq \#\} \ (= N_0(S) \cup N_1(S))$$

Thus, a glance at the 0 and 1 components of $z^4$ in the Construction Example above shows that $N_0(S) = \{3, 11\}$ and $N_1(S) = \{2, 6\}$. (In other words, these two sets of indexes respectively identify the variables $z_j$ such that $z_j = 0$ and such that $z_j = 1$ in every solution $z$ in $S$). Then $N_{01}(S) = \{2, 3, 6, 11\}$.

To identify the associated cardinalities of these sets, we define

$$|z|_v = |\{j \in N: z_j = v\}| \text{ for } v = 0 \text{ and } 1$$

$$|z|_v = |z|_0 + |z|_1 \text{ for } v = 01$$

and thus obtain

$$|N_v(S)| = |z|_v \text{ for } v = 0, 1 \text{ and } 01.$$

We use these relationships to characterize our algorithms for Problems $G_v^1(x(h))$ and $G_v^2(x(h))$, and thus to identify useful consistency information in the set of solutions $R$.

### 4.2.2. Constructive Algorithms for Problems $G_v^1(x(h))$ and $G_v^2(x(h))$

We restate Problems $G_v^1$ and $G_v^2$ by including the stipulation that $S$ contains a given solution $x$ in $R$, and using the alternative representation of Section 3, as follows. As before, let $L_v^1$ denote a chosen lower limit on the size of $N_v(S)$, hence on $|z|_v = \cap_o(S)$, and let and $L_v^2$ denote a chosen lower limit on the size of $S$, for $v = 0$ or $1$ or $01$. Then we state:

**Problem $G_v^1(x(h))$:** For a specified $x(h) \in R$, find a set $S$ with analog vector $z = \cap_o(S)$ to

$$(G_v^1(x(h))) \quad \begin{cases} max & |S| \\ s.t. & |z|_v \geq L_v^1 \\ & x(h) \in S \\ & S \subseteq R \end{cases}$$

**Problem $G_v^2(x(h))$:** For a specified $x(h) \in R$, find a set $S$ with analog vector $z = \cap_o(S)$ to

$$(G_v^2(x(h))) \quad \begin{cases} max & |z|_v \\ s.t. & |S| \geq L_v^2 \\ & x(h) \in S \\ & S \subseteq R \end{cases}$$

We find an approximate solution to $G_v^1(x(h))$ by identifying a maximal $S$ in $R$ satisfying the stated conditions and find an approximate solution to $G_v^2(x(h))$ in a similar manner. Recall $|z|_v = $ the number of components of $z$ with $z_j = v$, interpreting $v = 01$ to mean $v = 0$ or $1$.

These algorithms rely on the principle that increasing the size of $S$ can only decrease the value $|z|_v$ or leave it unchanged.

**Algorithm 1: for Problem $G_v^1(x(h))$ (Maximize $|S|$: for $x(h) \in S$ and $L_v^1 \leq |z|_v$)**

Let $z^k = \cap_o(S)$ denotes the analog vector of the current set $S$ at iteration $k$. (Note that $k = |S|$ since a single solution is added to $S$ at each iteration.)
Choose $x(h) \in R$; $S = \{x(h)\}$; $k = 1$; $z^1 = x(h)$; Done = False;
**While** Done = False and $S \neq R$ **do**
    Select a vector $x \in R\backslash S$ that maximizes $|z|_v$ for $z = z^k \cap_o x$, and denote the chosen vector
        by $x^*$. Hence $x^* = argmax\{|z^k \cap_o x|_v : x \in R\backslash S\}$;
    **If** $|z^*|_v \geq L_v^1$ **then** set $S = S \cup \{x^*\}$, $z^{k+1} = z^k \cap_o x^*$, $k = k + 1$;
    **Else** the method stops with the current $S$, Done = True;
**Endwhile**

**Algorithm 2: for Problem $G_v^2(x(h))$ (Maximize $|z|_v$: for $x(h) \in S$ and $L_v^2 \leq |S|$)**

At iteration $k = 1, 2, \ldots$, let $z^k = \cap_o(S)$ denote the analog vector of the current set $S$, where initially
Choose $x(h) \in R$; $S = \{ x(h)\}$; $k = 1$; $z^1 = x(h)$; Done = False;
**While** Done = False and $S \neq R$ **do**
    Select a vector $x \in R\backslash S$ that maximizes $|z|_v$ for $z = z^k \cap_o x$, and denote the chosen vector
       by $x^*$. Hence
    $x^* = argmax\{|z^k \cap_o x|_v: x \in R\backslash S\}$
    $S = S \cup \{x^*\}$, $z^{k+1} = z^k \cap_o x^*$, $k = k + 1$;
    If $L_v^2 = |S|$ ( $= k + 1$) **stop** with the current $S$.
**Endwhile**

Note this Algorithm outline is the same as for Problem $G_v^1(x(h))$ except for the last few instructions before "Continue next iteration." Specifically, the instructions

    **if** $|z^*|_v \geq L_v^1$ **then** set $S = S \cup \{x^*\}$, $z^{k+1} = z^k \cap_o x^*$, $k = k + 1$;
    **else** (1) is violated and the method stops with the current $S$, Done = True

are replaced by the instructions

    $S = S \cup \{x^*\}$, $z^{k+1} = z^k \cap_o x^*$, $k = k + 1$;
    If $L_v^2 = |S|$ ( $= k + 1$) **stop** with the current $S$.

### 4.2.3. Efficient Implementation of Constructive Algorithms

In this section, we describe an efficient implementation of constructive Algorithms 1 and 2 for problems $G_v^1$ and $G_v^2$ using set notation and the vector notation. The efficiency of these algorithms is based on the following observation.

**Observation 1:** Once $z_j$ receives the value # at iteration $k$, it continues to receive this value in all iterations with larger values of $k$. To exploit this, it is not necessary to compute the intersection vectors $z \cap_0 x(i)$ and $z \cap_0 x(i^*)$ for all components $j \in N$ but just over the set $N^* = \{j \in N : z_j \neq \#\} = \{j \in N : z_j \in \{0,1\}\}$.

The consequences of this observation are exploited in the determination of $x^* = argmax\{|z^k \cap_o x|_v : x \in R\backslash S\}$ and also in the update step.

**Observation 2:** We also observe another interesting relationship that follows from our preceding definitions. Let $S'$ and $S''$ be two subsets in $R$. We have

$$|N_v(S' \cup S'')| \leq Min(|N_v(S')|, |N_v(S'')|)$$

and

$$\text{if } S' \subseteq S'' \text{ or } S'' \subseteq S' \text{ then } |N_v(S' \cup S'')| = Min(|N_v(S')|, |N_v(S'')|).$$

In our constructive algorithm we will observe this relationship for the case where we want to enlarge a set $S$ by adding a solution $x$ to it to create the set $S \cup \{x\}$. Then

$$|N_v(S \cup \{x\})| \leq Min(|N_v(S)|, |N_v(\{x\})|)$$

The consequences of this observation are exploited in the initialization phase by excluding vectors $x(i) \in R$ such that $|x(i)|_v < L_v^1$ for $v \in \{0,1\}$ and also in the While loop, if $x(i)$ is not feasible for Problem $G_v^1$ and will never be feasible on later. In the pseudo-code, the set $E$ is used to save the excluded vectors.

In summary, the method goes through all $x(i) \in R \setminus (S \cup F)$ and picks the one $x(i^*)$ that gives the largest $n_v^*$ value for $n_v = N_v(S \cup x(i))$. This $x$ is added to $S$ if $n_v^* \geq L_v^1$, and the method terminates otherwise.

**Accelerated Algorithm 1: for Problem $G_v^1(x(h))$ (Maximize $|S|$: for $x(h) \in S$ and $L_v^1 \leq |z|_v$)**

    Choose $x(h) \in R$, $S = \{x(h)\}$, $F = \emptyset$, $z = x(h)$, $N^* = N_v(z)$, Done = False
    **For** each $x(i) \in R \setminus (S \cup F)$ **do** % This need only be executed for $v \in \{0,1\}$
        **If** $|x(i)|_v < L_v^1$ **then**
            $F = F \cup \{x(i)\}$
        **Endif**
    **Endfor**
    **While** Done = False and $S \cup F \neq R$ and $N^* \neq \emptyset$ **do**
        $n_v^* = -1$;
        **For** each $x(i) \in R \setminus (S \cup F)$ **do**
            $n_v = |\{j \in N^*: z_j \cap_0 x(i)_j = v\}|$
            **If** $n_v > n_v^*$ **then**
                $n_v^* = n_v$, $i^* = i$, $x_0^* = f(x(i^*))$
            **Endif**
            **If** $n_v < L_v^1$ **then**
                $F = F \cup \{x(i)\}$
            **Endif**
        **Endfor**
        **% Execute Update/Terminate Routine for Problem $G_v^1(x(h))$**
        **If** $n_v^* \geq L_v^1$ **then**
            $S = S \cup \{x(i^*)\}$
            **For** $j \in N^*$ **do** $z_j \cap_0 x(i)_j$
            $N^* = N^* \setminus \{j \in N^*: z_j = \#\}$
        **Else**
            Done = True
        **Endif**
    **Endwhile**

The detailed algorithm for Problem $G_v^2(x(h))$ employs a corresponding design, which can be inferred from the above and the description of the $G_v^2(x(h))$ algorithm in Section 4.2.2.

Appendix 1 shows how the foregoing algorithms can be equivalently expressed by reference to notation involving the index sets $N_v(x)$ in place of the vectors $z_v$ and gives a more fully detailed version of Algorithm 1 designed for greater efficiency.

Next, we describe advanced considerations for these algorithms and then in Section 6 give a numerical example of applying these algorithms.

The following illustration discloses additional advanced considerations that become apparent by numerical example.

## 5. Illustration
### 5.1. Algorithm 1 for Problem $G_v^1$

We illustrate Algorithm 1 for Problem $G_v^1$ by reference to a set $R$ containing 12 solutions, labeled $x(1)$ to $x(12)$. These solutions are organized in the Algorithm Example below so that the sets $S$ derived from each of the three objectives associated with $v = 0$, 1 and 01 can be illustrated simultaneously in the table for this example. Later comments show how the example can be applied to Algorithm 2, and to an adaptive method that may be viewed as an extension of both Algorithms 1 and 2.

It should be observed that our following illustration does not show the $z$ vectors corresponding to the successive sets $S$ generated for a given $v$ value. (Consequently, no "# components" appear in the vectors.) However, the $z$ vectors can readily be inferred by inspection based on our associated commentary and the fact that the size of $S$ for each case is small. This has an advantage of allowing the outcomes for different choices of $v$ to be considered in a single table. Specifically, the $|z|_v$ values listed immediately to the right of the solutions include all three cases for $v = 0$, 1 and 01. For emphasis, the entries for $|z|_v$ values that correspond to the goal of maximizing $|S|$ in each case are shown in bold face.

The Algorithm Example shown here is extended in Appendix 2 to identify the $z$ vectors for the cases $v = 0$ and $v = 1$, starting from the point where the table in the Algorithm Example leaves off.

The seed solutions for $v = 0$, 1 and 01 for this example are given by $x^1 = x(1)$, $x(4)$ and $x(6)$ respectively. The choices for $x^1$ correspond to those given in the Master Algorithm of Section 6. Thus $x^1 = x(1)$ for $v = 0$ because $x(1)$ is the solution containing the most components with $x_j = 0$, and similarly $x^1 = x(4)$ for $v = 1$ because $x(4)$ is the solution containing the most components with $x_j = 1$. Finally, $x^1 = x(6)$ for $v = 01$ because that the number of components with $x_j = 0$ and $x_j = 1$ are more nearly balanced in $x(6)$.

We first consider the case for $v = 0$ which appears first in the Algorithm Example. Here, $S$ is built in three iterations starting with $S = \{x(1)\}$ (since $x^1 = x(1)$). This yields a value for $|z|_0$ of 8 as shown in bold to the right of $x(1)$ (in the column under $v = 0$). The value $|z|_0 = 8$ for $S = \{x(1)\}$ is confirmed by counting the number of 0s in $x(1)$. The next iteration adds $x(2)$ to $S$ (as shown under "$S =$"), because $x(2)$ is the solution that maximizes $|z|_0$ when added to $S = \{x(1)\}$. This yields a value for $|z|_0$ of 6, as confirmed by counting the number of variables where $x_j = 0$ in both $x(1)$ and $x(2)$ (and as listed in the column under $v = 0$ for $x(2)$). Appendix 2 illustrates the basis for identifying which solution yields the largest $|z|_0$ value when added to $S$.

Finally, the third $S$ produced for $v = 0$ in the Algorithm Example shows that $x(3)$ is the best solution to join the set $S = \{x(1), x(2)\}$ when $v = 0$, by yielding $|z|_0 = 5$. This $|z|_0$ value may be similarly confirmed by counting the variables $x_j$ that equal 0 in all three solutions $x(1)$, $x(2)$ and $x(3)$. Since the lower bound $L_0^1 = 5$, the only way to expand $S$ further would be to find a solution to add to $S = \{x(1), x(2), x(3)\}$ that again yields $|z|_0 = 5$. However, as shown in Appendix 2, the best solution to add to the current $S$ yields $|z|_0 = 3$, and consequently the construction of $S$ for $v = 0$ ends here.

**Algorithm Example: Lower bounds $L_0^1 = 5$, $L_1^1 = 4$, $L_{01}^1 = 9$.**

| j = | Vectors in R | | | | | | | | | | | | $\|z\|_v = \|N_v(S)\|$ for v = | | | S = |
|---|---|---|---|---|---|---|---|---|---|---|---|---|---|---|---|---|
| | 1 | 2 | 3 | 4 | 5 | 6 | 7 | 8 | 9 | 10 | 11 | 12 | 0 | 1 | 01 | |
| x(1) | 1 | 1 | 0 | 1 | 0 | 0 | 0 | 0 | 1 | 0 | 0 | 0 | **8** | 4 | 12 | {x(1)} |
| x(2) | 0 | 0 | 1 | 1 | 1 | 0 | 0 | 0 | 1 | 0 | 0 | 0 | **6** | 2 | 8 | {x(1), x(2)} |
| x(3) | 1 | 1 | 1 | 0 | 1 | 0 | 1 | 0 | 1 | 0 | 0 | 0 | **5** | 1 | 6 | {x(1), x(2), x(3)} |
| Above generates S for v = 0 (with \|S\| = 3) | | | | | | | | | | | | | | | | |
| x(4) | 1 | 1 | 1 | 1 | 1 | 1 | 1 | 1 | 0 | 0 | 0 | 0 | 4 | **8** | 12 | {x(4)} |
| x(5) | 1 | 1 | 1 | 0 | 1 | 1 | 1 | 0 | 0 | 0 | 1 | 0 | 3 | **6** | 9 | {x(4), x(5)} |
| x(3)* | 1 | 1 | 1 | 0 | 1 | 0 | 1 | 0 | 1 | 0 | 0 | 0 | 2 | **5** | 7 | {x(4), x(5), x(3)*} |
| Above generates S for v = 1 (with \|S\| = 3) | | | | | | | | | | | | | | | | |
| x(6) | 0 | 1 | 0 | 1 | 0 | 1 | 0 | 1 | 0 | 1 | 0 | 1 | 6 | 6 | **12** | {x(6)} |
| x(7) | 0 | 1 | 0 | 1 | 0 | 1 | 0 | 1 | 0 | 0 | 1 | 1 | 5 | 5 | **10** | {x(6), x(7)} |
| x(8) | 0 | 1 | 0 | 0 | 0 | 1 | 0 | 1 | 0 | 1 | 1 | 1 | 5 | 4 | **9** | {x(6), x(7), x(8)} |
| Above generates S for v = 01 (with \|S\| = 3) | | | | | | | | | | | | | | | | |
| x(9) | 0 | 0 | 0 | 0 | 1 | 0 | 1 | 1 | 1 | 1 | 1 | 1 | 5 | **7** | 12 | {x(9)} |
| x(10) | 0 | 1 | 0 | 0 | 0 | 0 | 1 | 1 | 1 | 1 | 1 | 1 | 4 | **6** | 10 | {x(9), x(10)} |
| x(11) | 1 | 0 | 0 | 0 | 0 | 0 | 1 | 1 | 1 | 1 | 0 | 1 | 3 | **5** | 8 | {x(9), x(10), x(11)} |
| x(12) | 0 | 0 | 0 | 0 | 1 | 0 | 1 | 1 | 1 | 1 | 0 | 1 | 3 | **5** | 8 | {x(9), x(10), x(11), x(12)} |
| Above generates *different* S for v = 1 (with \|S\| = 4) | | | | | | | | | | | | | | | | |

We next consider the case where $v = 1$ and $x^1 = x(4)$. A similar progression ensues by adding first $x(5)$ and then $x(3)$, to yield a succession of sets $S = \{x(4)\}$, $S = \{x(4), x(5)\}$ and $S = \{x(4), x(5), x(3)\}$ with corresponding $|z|_1$ values given by $|z|_1 = 8$, 6 and 5. (By chance, this is the same succession of values produced for $|z|_0$ in the $v = 0$ case previously illustrated.) The solution $x(3)$ is duplicated in the Algorithm Example (and marked the second time with "*") to allow the basis for adding $x(3)$ as the third solution for the $v = 1$ case to be seen more easily. Here we stop after adding $x(3)$ with $|z|_1 = 5$. Although the lower bound for $v = 1$ is $L_1^1 = 4$, Appendix 2 shows that the next solution to be added to S yields $|z|_1 = 2$.

### 5.2. ε-constraint method

We show in the following table the execution of ε-Constraint algorithm for $G_v$ problem for $p = 1$ on the example described in Section 6 with $n = r = 12$.

| iter | $L_v^2$ | \|S*\| | \|$N_v(S^*)$\| | S* | $N_v(S^*)$ |
|---|---|---|---|---|---|
| | | | | v = 0, \|F\|= 7 | |
| 1 | 0 | 12 | 0 | R | ∅ |
| 2 | 1 | 8 | 1 | {x(1), x(6), x(7), x(8), x(9), x(10), x(11), x(12)} | {3} |
| 3 | 2 | 6 | 2 | {x(1), x(6), x(7), x(8), x(10), x(11)} | {3, 5} |
| 4 | 3 | 4 | 3 | {x(1), x(2), x(3), x(4)} | {10, 11, 12} |
| 5 | 4 | 3 | 5 | {x(1), x(2), x(3)} | {6, 8, 10, 11, 12} |
| 6 | 6 | 2 | 6 | {x(1), x(2)} | {6, 7, 8, 10, 11, 12} |
| 7 | 7 | 1 | 8 | {x(1)} | {3, 5, 6, 7, 8, 10, 11, 12} |
| $v = 1, |F| = 8$ | | | | | |
| 1 | 0 | 12 | 0 | R | ∅ |
| 2 | 1 | 8 | 1 | {x(1), x(3), x(4), x(5), x(6), x(7), x(8), x(10)} | {2} |
| 3 | 2 | 7 | 2 | {x(6), x(7), x(8), x(9), x(10), x(11), x(12)} | {8, 12} |
| 4 | 3 | 6 | 3 | {x(6), x(8), x(9), x(10), x(11), x(12)} | {8, 10, 12} |
| 5 | 4 | 4 | 5 | {x(9), x(10), x(11), x(12)} | {7, 8, 9, 10, 12} |
| 6 | 6 | 2 | 6 | {x(9), x(12)} | {5, 7, 8, 9, 10, 12} |
| 7 | 7 | 1 | 7 | {x(9)} | {5, 7, 8, 9, 10, 11, 12} |
| 8 | 8 | 1 | 8 | {x(4)} | {1, 2, 3, 4, 5, 6, 7, 8} |
| $v = 01, |F| = 11$ | | | | | |
| 1 | 0 | 12 | 0 | R | ∅ |
| 2 | 1 | 8 | 1 | {x(1), x(3), x(4), x(5), x(6), x(7), x(8), x(10)} | {2} |
| 3 | 2 | 7 | 3 | {x(6), x(7), x(8), x(9), x(10), x(11), x(12)} | {3, 8, 12} |
| 4 | 4 | 6 | 4 | {x(6), x(8), x(9), x(10), x(11), x(12)} | {3, 8, 10, 12} |
| 5 | 5 | 5 | 5 | {x(8), x(9), x(10), x(11), x(12)} | {3, 4, 8, 10, 12} |
| 6 | 6 | 4 | 6 | {x(6), x(7), x(8), x(10)} | {1, 2, 3, 5, 8, 12} |
| 7 | 7 | 4 | 8 | {x(9), x(10), x(11), x(12)} | {3, 4, 6, 7, 8, 9, 10, 12} |
| 8 | 9 | 3 | 9 | {x(6), x(7), x(8)} | {1, 2, 3, 5, 6, 7, 8, 9, 12} |
| 9 | 10 | 2 | 10 | {x(6), x(8)} | {1, 2, 3, 5, 6, 7, 8, 9, 10, 12} |
| 10 | 11 | 2 | 11 | {x(9), x(12)} | {1, 2, 3, 4, 5, 6, 7, 8, 9, 10, 12} |
| 11 | 12 | 1 | 12 | {x(11)} | {1, 2, 3, 4, 5, 6, 7, 8, 9, 10, 11, 12} |

## 6. Advanced Considerations

The determination of the set $S$ by Algorithms 1 and 2 for problems $G_v^1$ and $G_v^2$ can be modified in several ways. First, we consider how these algorithms can be extended by applying them to different choices of the starting vector $x^1$ to generate more than one set $S$.

One possibility for generating multiple sets $S$ from different starting points $x^1$ is to select a new $x^1$ to be the last solution added to create a previous $S$. Although ideally it should be possible to generate the same $S$ by starting from any of its component solutions, this may not happen because Algorithms 1 and 2 are approximation methods.[1] Starting from the last solution added to $S$ gives an increased chance to discover variations caused by the inexact nature of these algorithms. (Such a process could be continued, for example, until a final $S$ duplicates a previous $S$, not necessarily the immediate predecessor.) However, to avoid unnecessary generation of sets that may not differ substantially from each other, it seems worthwhile to select new points $x^1$ that lie outside sets previously generated. We formalize this in the following algorithm that employs the algorithms for problems $G_v^1$ and $G_v^2$ as subroutines.

---

[1] This may also happen in the presence of alternative optima, where different ones may be encountered by selecting different starting points.

We make use of an upper limit $U$ on the number $n_S$ of sets $S$ generated and a lower limit $L$ on the size of $S$ so that when either $n_S$ reaches $U$ or a set $S$ generated fails to satisfy $|S| \geq L$ the generation of new sets $S$ terminates. ($L$ has a function similar to that of $L_v^2$ in Algorithm 2 for $G_v^2$, in this case functioning as an ultimate limit.)

**Master Algorithm:**

*Initialization:* $n_S = 0$, $C = \emptyset$.

**Do{**

> Select $x^1 \in R \setminus C$ that maximizes $|x|_v$ (the number of components $x_j$ that equal $v$) for
> 
> > $v = 0$ or 1, or that maximizes $| |x|_1 - |x|_0 |$ for $v = 01$;
> 
> Execute Algorithm 1 or Algorithm 2 to generate a set $S$;
> 
> $n_S = n_S + 1$
> 
> $C = S \cup C$;

**} While** $|S| \geq L$ and $n_S < U$;

**Remark 4**: If the convention $v = 01$ is coded by $v = ½$, the instruction

> Select $x^1 \in R \setminus C$ that maximizes $|x|_v$ (the number of components $x_j$ that equal $v$) for
> 
> > $v = 0$ or 1, or that maximizes $| |x|_1 - |x|_0 |$ for $v = 01$;

can be replaced by

$$x^1 = argmax\{|(1-v)|x|_0 - v|x|_1| \text{ for } x \in R\setminus C\};$$

or equivalently

$$x^1 = argmax\{|(1-v)e\bar{x} - vex|: x \in R\setminus C\}.$$

We now elaborate this description of the Master Algorithm with several important considerations.

### 6.1. Quasi-clusters

We note the interesting fact that the sets $S$ for $v = 0$ and $v = 1$ can have overlaps, here consisting of the solution $x(3)$. We call sets having such overlaps *quasi-clusters* to differentiate them from customary clusters which are not permitted to have overlaps. The existence of quasi-clusters in the present setting, where the sets are generated from different starting solutions and constructed according to different goals, invites exploration to see whether such overlaps are common for problems commonly encountered, and to identify the significance of such overlaps when they occur. (If the values chosen for $L_v^1$ are small enough, however, it is clear that a large number of overlaps may be expected.)

The case for $v = 01$ in the preceding example generates a set $S$ that does not overlap with those for other $v$ values and produces sets $S$ with larger values of $|z|_v$. It should not be surprising that $v = 01$ enables a larger number of variables to be fixed than in the cases for $v = 0$ and $v = 1$, though the relevance of this undoubtedly depends on the problem setting. In the present example, we have set a lower bound $L^1_{01} = 9$ for $v = 01$, which is somewhat larger than the lower bounds chosen for $v = 0$ and $v = 1$.

Finally, the Algorithm Example shows the result of picking a different starting solution $x^1$ for $v = 1$. This second choice for $x^1$ follows the proposal of the Master Algorithm in Section 4, which selects a new starting solution to be one that contains a largest number of components $x_j = v$, subject to the condition that this solution does not lie in a set $S$ previously generated for $v$. Thus, $x(9)$, which does not belong to the previous set $S$ for $v = 1$, and which contains 7 variables with $x_j = 1$, is the new choice for $x^1$.

Although $S = \{x(9)\}$ does not yield $|z|_1$ as large as produced by the earlier choice that gives $S = \{x(4)\}$, the ultimate set $S = \{x(9), x(10), x(11), x(12)\}$ has a larger size ($|S| = 4$) than in the previous instance. The two sets $S$ generated for $v = 1$ in this example are disjoint, suggesting that the two cases have a meaningful difference from the standpoint of forming clusters.

This illustrates another interesting phenomenon. Suppose we use a simple frequency count of the number of times that the assignments $x_j = v$ appear in the solutions $x(1)$ to $x(12)$ as a basis for assigning probabilities for creating new solutions, and consider the case for $x_j = 1$. The resulting frequencies are as follows.

| $j$ | 1 | 2 | 3 | 4 | 5 | 6 | 7 | 8 | 9 | 10 | 11 | 12 |
|---|---|---|---|---|---|---|---|---|---|---|---|---|
| Frequency = $\sum_{i=1}^{r=12} x(i)_j$ | 5 | 8 | 4 | 5 | 6 | 5 | 7 | 8 | 7 | 6 | 5 | 7 |

Grouped by frequency, $x_2$ and $x_8$ are the most attractive for setting $x_j = 1$, followed by $x_7$, $x_9$ and $x_{12}$, while $x_1$, $x_3$, $x_4$, $x_5$ and $x_{11}$ are the least attractive. However, $x_2 = 1$ does not show up in any solutions where $x_8$ receives the value 1, but instead is associated with the less attractive variables $x_1$, $x_3$, and $x_5$ (in the set $S = \{x(4), x(5), x(3)\}$ generated above). This shows that a measure of "attractiveness" by frequency counts can be misleading and fail to capture associations between variables. Interestingly, however, the example suggests that we may use the appearance of a variable in more than one set $S$ based on the same $v = 0$ or 1 as an indication of being strongly determined without the computational expense otherwise required to identify such variables. Here, $x_7 = 1$ appears in both of the sets $S$ illustrated in the Algorithm Example for $v = 1$, and hence may be viewed as strongly determined (in spite of the fact that $x_2$ and $x_8$ have higher frequency counts). In short, a "higher order" frequency count, which identifies the number of occurrences of $x_j = v$ in the sets $S$ generated by Algorithm 1 or 2 for a given $v$, gives a useful basis for identifying variables that may be considered strongly determined.

As alluded to in earlier comments, a further elaboration of the underlying algorithmic processes for this example is provided in Appendix 2 by examining possible "next $S$ sets" starting from the final sets in this example. An inference from Appendix 2 is worth emphasizing here. The information from the tables in this appendix shows that the best remaining evaluation for $v = 0$ yields $|z|_0 = 3$ as remarked above. If $L^1_0$ was selected to be 3 instead of 4 (as currently stipulated), the next solution added to $S$ for $v = 0$ after $x(3)$ in the Algorithm Example would be $x(4)$ or $x(5)$. However, this drop from $|z|_0 = 5$ to $|z|_0 = 3$ represents a decline in $|z|_0$ from $.42n$ to $.25n$. An even greater decline occurs for the case $v = 1$, where $|z|_1 = 5$ drops to $|z|_1 = 2$. Consequently, this suggests that the algorithm might be modified to adaptively monitor $|z|_v$ to allow the method to stop before allowing a significant decline in $|z|_v$ by adding one more solution to $S$. The structure of the algorithm is congenial to making such a modification.

## 7. Computational Results

The proposed algorithms were implemented in C++ and run on a PC computer with a 2.9 Ghz processor. The MIPs were solved using CPLEX 12.6.1. To perform the computational experiments, we note that an instance of the problem is characterized by an input binary matrix $A(r, n)$ where $r$ is the number of rows and $n$ is the number of columns (see Section 3.1). Several sources exist in the literature where these matrices can be obtained from real-world data. In this paper, the binary matrices are generated randomly using a simple uniform distribution provided by the Microsoft Visual Studio. Specifically, for a given dimension (i.e. $n$ and $r$ are fixed), a component $a_{ij}$ of the matrix $A$, is generated randomly in $\{0, 1\}$. For each fixed dimension $(n, r)$, 10 instances are generated randomly.

We compare the performance of the proposed integer program $IP_v^1(x(h))$ with the constructive heuristic and its accelerated version. The maximum size of $n$ x $r$ is fixed to 500 x 500 or 200 x 20000 according the memory capacity of the computer used. The computation time for each integer program $IP_v^1(x(h))$ is limited to 3600 seconds. For some hard instances, we found that CPLEX crashed with an "Out of memory" message before reaching the time limit. In such cases we reduced the time limit to lie between 100 and 360 seconds in order to obtain a feasible solution for comparison.

For all computational tests we choose the target solution $x(h)$ as follows:

$$x(h) = \begin{cases} argmax\{|N_v(x(i))|: i \in R\} & if\ v \in \{0,1\} \\ argmax\{\left||N_0(x(i))| - |N_1(x(i))|\right|: i \in R\} & if\ v = 01 \end{cases}$$

For each instance with fixed dimension $n$ and $r$, the lower bound is chosen by setting:

$$L_v^1 = |N_v(R)| + \alpha(|N_v(x(h))| - |N_v(R)|)$$

where $\alpha \in \{0.2, 0.4, 0.6, 0.8\}$.

The legend of the columns of the tables uses the following conventions:
- CPLEX refers to the MIP solver of CPLEX used to solve $IP_v^1(x(h))$.
- H corresponds to the constructive heuristic described in Algorithm 1
- AH corresponds to the accelerated version of the constructive heuristic described in Algorithm 2.
- $|S^*|$ is the size of the best set of rows generated by the corresponding approach.
- $|N_v(S^*)|$ is the size of the intersect set corresponding to $S^*$.
- CPU is the computation time in seconds needed by the procedure.
- #Opt is the number of instances solved optimally by CPLEX.
- #E1 is the number of best solutions generated by the constructive heuristics with the same first objective $|S^*|$ as used by CPLEX (i.e. MIP($|S^*|$) = H($|S^*|$)).

- #E2 is the number of best solutions generated by the constructive heuristics with the same first objective value $|S^*|$ as generated by CPLEX and the second objective value $|N_v(S^*)|$ is strictly greater than the one generated by CPLEX (i.e. MIP($|S^*|$) = H($|S^*|$) and MIP($|N_v(S^*)|$) < H($|N_v(S^*)|$)). Hence #E2 identifies the number of solutions generated by the constructive heuristics that dominate those obtained by CPLEX.

Each row is an average of the values computed over various instances and the last row corresponds to the average of all instances.

Tables 1.1, 1.2 and 1.3 compare the results of the MIP solver (CPLEX), the constructive heuristic (H) and its accelerated version (AH) for value $v = 0$, 1 and 01 respectively, on small-sized instances with $n = 125$, $r \in \{12, 25, 62\}$ and $\alpha \in \{0.2, 0.4, 0.6, 0.8\}$. Each line corresponds to the average of 50 instances (10 randomly generated instances with 5 different densities). The computational time for the heuristics H and AH are not reported since they are negligible (i.e. less than 0.0005 second). Also, the solution objective values $|S^*|$ and $|N_v(S^*)|$ for H and AH are the same in all cases and hence are reported under the single heading "H - AH".

For $v = 0$ (see Table 1.1), most of the 600 instances are solved optimally except for 7 instances for $r = 62$ and $\alpha \in \{0.2, 0.4\}$ (i.e. six instances are not solved optimally due to the memory limitation "Out of Memory" error and one instance is not solved optimally within the time limit of one hour). Table 1.1 shows also that the constructive heuristic reaches an optimal value for 84% of the instances and provides solutions that dominate those provided by CPLEX in 49% of the instances. For $v = 1$ (see Table 1.2), all 600 instances are solved optimally by CPLEX. The constructive heuristic reaches an optimal value for 97% of the instances and provides solutions that dominate those provided by CPLEX in 24% of the instances. For $v = 01$ (see Table 1.3), CPLEX solves optimally 98% of the instances (11 instances are not solved optimally due to the memory limitation "Out of Memory" error) while the constructive heuristic reaches optimal value for 87% of the instances and provides solutions that dominate those provided by CPLEX in 43% of the instances. Note also that regarding the CPU time, the hardest instances correspond to instances with the parameters ($v \in \{0,1,01\}$, $r = 62$ and $\alpha \in \{0.2, 0.4\}$).

|   |   | MIP | | | H - AH | | | | |
|---|---|---|---|---|---|---|---|---|---|
| r | α | $\|S^*\|$ | $\|N_v(S^*)\|$ | CPU | $\|S^*\|$ | $\|N_v(S^*)\|$ | #Opt | #E1 | #E2 |
| 12 | 0.2 | 6.36 | 26.60 | 0.069 | 6.30 | 29.00 | 50 | 47 | 31 |
| 12 | 0.4 | 4.24 | 45.70 | 0.064 | 4.20 | 49.00 | 50 | 47 | 37 |
| 12 | 0.6 | 2.66 | 68.20 | 0.030 | 2.60 | 71.00 | 50 | 47 | 24 |
| 12 | 0.8 | 1.60 | 87.80 | 0.005 | 1.60 | 88.00 | 50 | 50 | 6 |
| 25 | 0.2 | 9.52 | 20.60 | 1.200 | 9.10 | 23.00 | 50 | 30 | 19 |
| 25 | 0.4 | 5.66 | 41.30 | 0.340 | 5.50 | 46.00 | 50 | 43 | 36 |
| 25 | 0.6 | 3.52 | 63.10 | 0.120 | 3.50 | 65.00 | 50 | 48 | 32 |
| 25 | 0.8 | 1.78 | 87.70 | 0.027 | 1.80 | 89.00 | 50 | 49 | 16 |
| 62 | 0.2 | 13.80 | 19.30 | 450.000 | 13.00 | 20.00 | 45 | 18 | 12 |
| 62 | 0.4 | 7.48 | 39.50 | 250.000 | 7.00 | 44.00 | 48 | 30 | 27 |
| 62 | 0.6 | 4.32 | 60.10 | 8.300 | 4.20 | 63.00 | 50 | 47 | 34 |
| 62 | 0.8 | 2.06 | 88.00 | 0.220 | 2.10 | 89.00 | 50 | 50 | 17 |
| Average | | 5.25 | 53.99 | 59.20 | 5.08 | 56.33 | 49.42 | 42.17 | 24.25 |

Table 1.1: Comparison of MIP, Heuristic and Accelerated Heuristic for $n = 125$ and $v = 0$

| $R$ | $\alpha$ | MIP | | | H - AH | | #Opt | #E1 | #E2 |
|---|---|---|---|---|---|---|---|---|---|
| | | $|S^*|$ | $|N_v(S^*)|$ | CPU | $|S^*|$ | $|N_v(S^*)|$ | | | |
| 12 | 0.2 | 2.64 | 10.20 | 0.056 | 2.60 | 12.00 | 50 | 49 | 27 |
| 12 | 0.4 | 1.52 | 28.50 | 0.033 | 1.50 | 30.00 | 50 | 50 | 16 |
| 12 | 0.6 | 1.04 | 44.10 | 0.003 | 1.00 | 44.00 | 50 | 50 | 1 |
| 12 | 0.8 | 1.00 | 45.30 | 0.002 | 1.00 | 45.00 | 50 | 50 | 0 |
| 25 | 0.2 | 2.96 | 9.62 | 0.120 | 2.80 | 12.00 | 50 | 44 | 34 |
| 25 | 0.4 | 1.74 | 26.00 | 0.072 | 1.70 | 28.00 | 50 | 49 | 16 |
| 25 | 0.6 | 1.12 | 43.20 | 0.004 | 1.10 | 43.00 | 50 | 50 | 1 |
| 25 | 0.8 | 1.00 | 46.60 | 0.003 | 1.00 | 47.00 | 50 | 50 | 0 |
| 62 | 0.2 | 3.50 | 9.64 | 1.500 | 3.30 | 11.00 | 50 | 42 | 22 |
| 62 | 0.4 | 1.86 | 25.40 | 0.460 | 1.80 | 28.00 | 50 | 47 | 21 |
| 62 | 0.6 | 1.18 | 43.30 | 0.008 | 1.20 | 43.00 | 50 | 50 | 5 |
| 62 | 0.8 | 1.00 | 48.70 | 0.006 | 1.00 | 49.00 | 50 | 50 | 0 |
| Average | | 1.71 | 31.71 | 0.19 | 1.67 | 32.67 | 50.00 | 48.42 | 11.92 |

Table 1.2: Comparison of MIP, Heuristic and Accelerated Heuristic for $n = 125$ and $v = 1$

| $r$ | $\alpha$ | MIP | | | H - AH | | #Opt | #E1 | #E2 |
|---|---|---|---|---|---|---|---|---|---|
| | | $|S^*|$ | $|N_v(S^*)|$ | CPU | $|S^*|$ | $|N_v(S^*)|$ | | | |
| 12 | 0.2 | 5.78 | 33.60 | 0.088 | 5.70 | 36.00 | 50 | 46 | 32 |
| 12 | 0.4 | 3.66 | 61.90 | 0.100 | 3.60 | 67.00 | 50 | 47 | 40 |
| 12 | 0.6 | 2.24 | 95.70 | 0.039 | 2.20 | 97.00 | 50 | 49 | 19 |
| 12 | 0.8 | 1.20 | 122.00 | 0.005 | 1.20 | 120.00 | 50 | 50 | 3 |
| 25 | 0.2 | 8.64 | 26.60 | 1.700 | 8.30 | 30.00 | 50 | 35 | 26 |
| 25 | 0.4 | 5.06 | 54.50 | 0.370 | 5.00 | 59.00 | 50 | 45 | 33 |
| 25 | 0.6 | 2.90 | 86.50 | 0.120 | 2.90 | 89.00 | 50 | 50 | 29 |
| 25 | 0.8 | 1.46 | 119.00 | 0.025 | 1.50 | 120.00 | 50 | 50 | 3 |
| 62 | 0.2 | 12.40 | 25.20 | 830.000 | 11.00 | 27.00 | 41 | 23 | 15 |
| 62 | 0.4 | 6.60 | 52.30 | 97.000 | 6.30 | 56.00 | 48 | 37 | 26 |
| 62 | 0.6 | 3.72 | 78.60 | 7.100 | 3.60 | 83.00 | 50 | 43 | 25 |
| 62 | 0.8 | 1.76 | 116.00 | 0.150 | 1.70 | 120.00 | 50 | 48 | 8 |
| Average | | 4.62 | 72.66 | 78.06 | 4.42 | 75.33 | 49.08 | 43.58 | 21.58 |

Table 1.3: Comparison of MIP, Heuristic and Accelerated Heuristic for $n = 125$ and $v = 01$

Tables 2.1, 2.2 and 2.3 compare the results of the MIP solver (CPLEX), the constructive heuristic (H) and its accelerated version (AH) for value $v = 0, 1$ and $01$ respectively, on meduim-sized instances with $n \in \{125, 250, 500\}$, $r = \lfloor \beta n \rfloor$ with $\beta \in \{0.1, 0.2, 0.5, 1\}$ and $\alpha \in \{0.2, 0.4, 0.6, 0.8\}$. Each line corresponds to the average of 10 randomly generated instances for a fixed value of $n$, $r$ and $\alpha$. Once again, H and AH obtain the same objective values but now their solution times differ enough in some cases to be reported separately.

For $v = 0$ (see Table 2.1), 86% of the instances are solved optimally except for 68 instances over 480 (i.e. only one instance is not solved optimally in the time limit of one hour and the other 67 instances are not solved optimally due to the memory limitation "Out of Memory" error). Table 2.1 shows also that the constructive heuristic reaches optimal value for 90% of the instances and provides solutions that dominates those provided by CPLEX in 36% of the instances. For $v = 1$ (see Table 2.2), 87% of the instances are solved optimally except for 63 instances over 480 (i.e. the other 63 instances are not solved optimally due to the memory limitation "Out of Memory" error). Table 1.2 shows that the constructive heuristic reaches an optimal value for 93% of the instances and provides solutions that dominate those provided by CPLEX in 40% of the instances. For $v = 01$ (see Table 2.3), 84% of the instances are solved optimally except for 79 instances over 480 (i.e. 10 instances are not solved optimally in the time limit of one hour and the other 69 instances are not solved optimally due to the memory limitation "Out of Memory" error). The constructive heuristic reaches an optimal value for 95% of the instances and provides solutions that dominate those provided by CPLEX in 43% of the instances. Note also that regarding the CPU time consumed by CPLEX, the hardest instances correspond to instances with the parameter $\alpha \in \{0.2, 0.4\}$.

| | | | MIP | | | H | | | AH | | | |
|---|---|---|---|---|---|---|---|---|---|---|---|---|
| $n$ | $r$ | $\alpha$ | $\|S^*\|$ | $\|N_v(S^*)\|$ | CPU | $\|S^*\|$ | $\|N_v(S^*)\|$ | CPU | CPU | #Opt | #E1 | #E2 |
| 125 | 12 | 0.2 | 4.0 | 14.4 | 0.125 | 3.9 | 15.8 | 0.000 | 0.000 | 10 | 9 | 3 |
| 125 | 12 | 0.4 | 2.0 | 37.6 | 0.135 | 2.0 | 41.8 | 0.000 | 0.000 | 10 | 10 | 8 |
| 125 | 12 | 0.6 | 1.3 | 63.1 | 0.004 | 1.3 | 63.1 | 0.000 | 0.000 | 10 | 10 | 0 |
| 125 | 12 | 0.8 | 1.0 | 71.9 | 0.006 | 1.0 | 71.9 | 0.000 | 0.000 | 10 | 10 | 0 |
| 125 | 25 | 0.2 | 4.4 | 14.5 | 0.401 | 4.1 | 17.1 | 0.000 | 0.000 | 10 | 7 | 6 |
| 125 | 25 | 0.4 | 2.5 | 33.8 | 0.180 | 2.5 | 36.7 | 0.000 | 0.000 | 10 | 10 | 6 |
| 125 | 25 | 0.6 | 1.7 | 53.7 | 0.006 | 1.7 | 54.4 | 0.000 | 0.000 | 10 | 10 | 3 |
| 125 | 25 | 0.8 | 1.0 | 73.9 | 0.003 | 1.0 | 73.9 | 0.000 | 0.000 | 10 | 10 | 0 |
| 125 | 62 | 0.2 | 5.0 | 14.8 | 5.610 | 4.6 | 17.0 | 0.000 | 0.000 | 10 | 6 | 4 |
| 125 | 62 | 0.4 | 2.7 | 33.8 | 1.707 | 2.6 | 37.2 | 0.000 | 0.000 | 10 | 9 | 2 |
| 125 | 62 | 0.6 | 2.0 | 45.3 | 0.015 | 2.0 | 46.2 | 0.000 | 0.000 | 10 | 10 | 6 |
| 125 | 62 | 0.8 | 1.0 | 75.1 | 0.009 | 1.0 | 75.1 | 0.001 | 0.000 | 10 | 10 | 0 |
| 125 | 125 | 0.2 | 5.5 | 15.0 | 66.747 | 5.3 | 16.3 | 0.001 | 0.000 | 10 | 8 | 5 |
| 125 | 125 | 0.4 | 3.0 | 30.7 | 7.153 | 3.0 | 32.9 | 0.000 | 0.000 | 10 | 10 | 9 |
| 125 | 125 | 0.6 | 2.0 | 47.6 | 0.072 | 2.0 | 49.6 | 0.000 | 0.000 | 10 | 10 | 7 |
| 125 | 125 | 0.8 | 1.0 | 76.6 | 0.014 | 1.0 | 76.6 | 0.000 | 0.000 | 10 | 10 | 0 |
| 250 | 25 | 0.2 | 3.9 | 28.8 | 0.707 | 3.9 | 33.1 | 0.000 | 0.000 | 10 | 10 | 10 |
| 250 | 25 | 0.4 | 2.0 | 73.6 | 0.279 | 2.0 | 82.3 | 0.000 | 0.000 | 10 | 10 | 9 |
| 250 | 25 | 0.6 | 1.3 | 124.1 | 0.006 | 1.3 | 124.1 | 0.000 | 0.000 | 10 | 10 | 0 |
| 250 | 25 | 0.8 | 1.0 | 141.1 | 0.006 | 1.0 | 141.1 | 0.000 | 0.000 | 10 | 10 | 0 |
| 250 | 50 | 0.2 | 4.0 | 27.9 | 9.831 | 4.0 | 33.9 | 0.000 | 0.000 | 10 | 10 | 10 |
| 250 | 50 | 0.4 | 2.0 | 72.9 | 2.329 | 2.0 | 84.3 | 0.002 | 0.000 | 10 | 10 | 10 |
| 250 | 50 | 0.6 | 1.7 | 102.2 | 0.013 | 1.7 | 102.2 | 0.000 | 0.000 | 10 | 10 | 0 |
| 250 | 50 | 0.8 | 1.0 | 141.6 | 0.008 | 1.0 | 141.6 | 0.000 | 0.000 | 10 | 10 | 0 |
| 250 | 125 | 0.2 | 4.4 | 29.0 | 810.670 | 4.0 | 37.1 | 0.000 | 0.000 | 9 | 6 | 6 |
| 250 | 125 | 0.4 | 2.5 | 68.4 | 32.983 | 2.4 | 76.3 | 0.000 | 0.000 | 10 | 9 | 4 |
| 250 | 125 | 0.6 | 1.6 | 113.1 | 0.021 | 1.6 | 113.6 | 0.000 | 0.000 | 10 | 10 | 1 |
| 250 | 125 | 0.8 | 1.0 | 146.8 | 0.019 | 1.0 | 146.8 | 0.000 | 0.000 | 10 | 10 | 0 |

| n | r | α | MIP |S*| | MIP |N_v(S*)| | MIP CPU | H |S*| | H |N_v(S*)| | H CPU | AH CPU | #Opt | #E1 | #E2 |
|---|---|---|---|---|---|---|---|---|---|---|---|---|
| 250 | 250 | 0.2 | 4.0 | 29.3 | 252.044 | 4.5 | 33.9 | 0.002 | 0.001 | 0 | 5 | 5 |
| 250 | 250 | 0.4 | 2.6 | 64.7 | 268.853 | 2.5 | 74.7 | 0.001 | 0.001 | 3 | 9 | 5 |
| 250 | 250 | 0.6 | 1.9 | 95.3 | 0.054 | 1.9 | 96.2 | 0.002 | 0.000 | 10 | 10 | 3 |
| 250 | 250 | 0.8 | 1.0 | 146.8 | 0.033 | 1.0 | 146.8 | 0.002 | 0.000 | 10 | 10 | 0 |
| 500 | 50 | 0.2 | 4.0 | 54.9 | 41.723 | 3.5 | 73.3 | 0.001 | 0.000 | 10 | 5 | 2 |
| 500 | 50 | 0.4 | 2.0 | 133.8 | 5.237 | 2.0 | 156.4 | 0.002 | 0.000 | 10 | 10 | 10 |
| 500 | 50 | 0.6 | 1.1 | 263.7 | 0.020 | 1.1 | 263.7 | 0.000 | 0.000 | 10 | 10 | 0 |
| 500 | 50 | 0.8 | 1.0 | 275.0 | 0.019 | 1.0 | 275.0 | 0.001 | 0.000 | 10 | 10 | 0 |
| 500 | 100 | 0.2 | 3.6 | 59.3 | 166.053 | 4.0 | 57.5 | 0.004 | 0.001 | 0 | 6 | 6 |
| 500 | 100 | 0.4 | 2.0 | 137.0 | 142.226 | 2.0 | 160.4 | 0.001 | 0.000 | 10 | 10 | 10 |
| 500 | 100 | 0.6 | 1.0 | 278.1 | 0.027 | 1.0 | 278.1 | 0.000 | 0.000 | 10 | 10 | 0 |
| 500 | 100 | 0.8 | 1.0 | 278.1 | 0.026 | 1.0 | 278.1 | 0.000 | 0.000 | 10 | 10 | 0 |
| 500 | 250 | 0.2 | 3.2 | 65.0 | 100.035 | 4.0 | 63.5 | 0.003 | 0.003 | 0 | 2 | 2 |
| 500 | 250 | 0.4 | 2.0 | 137.0 | 100.032 | 2.0 | 166.6 | 0.002 | 0.001 | 0 | 10 | 10 |
| 500 | 250 | 0.6 | 1.5 | 225.9 | 0.054 | 1.5 | 225.9 | 0.001 | 0.000 | 10 | 10 | 0 |
| 500 | 250 | 0.8 | 1.0 | 282.1 | 0.052 | 1.0 | 282.1 | 0.003 | 0.001 | 10 | 10 | 0 |
| 500 | 500 | 0.2 | 3.1 | 72.5 | 120.069 | 4.0 | 64.8 | 0.007 | 0.003 | 0 | 1 | 1 |
| 500 | 500 | 0.4 | 2.0 | 144.1 | 120.185 | 2.0 | 166.0 | 0.008 | 0.003 | 0 | 10 | 10 |
| 500 | 500 | 0.6 | 1.1 | 271.8 | 4.323 | 1.1 | 271.8 | 0.004 | 0.002 | 10 | 10 | 0 |
| 500 | 500 | 0.8 | 1.0 | 283.0 | 3.417 | 1.0 | 283.0 | 0.002 | 0.001 | 10 | 10 | 0 |
| **Average** | | | **2.2** | **105.6** | **47.157** | **2.2** | **109.6** | **0.001** | **0.000** | **8.6** | **9.0** | **3.6** |

Table 2.1: Comparison of MIP, Heuristic and Accelerated Heuristic for $v = 0$

| | | | MIP | | | H | | | AH | | | |
|---|---|---|---|---|---|---|---|---|---|---|---|---|
| n | r | α | |S*| | $|N_v(S^*)|$ | CPU | |S*| | $|N_v(S^*)|$ | CPU | CPU | #Opt | #E1 | #E2 |
| 125 | 12 | 0.2 | 4.0 | 14.3 | 0.165 | 3.7 | 17.0 | 0.000 | 0.000 | 10 | 7 | 2 |
| 125 | 12 | 0.4 | 2.0 | 37.2 | 0.134 | 2.0 | 41.3 | 0.000 | 0.001 | 10 | 10 | 6 |
| 125 | 12 | 0.6 | 1.3 | 63.3 | 0.002 | 1.3 | 63.4 | 0.000 | 0.000 | 10 | 10 | 1 |
| 125 | 12 | 0.8 | 1.0 | 71.6 | 0.002 | 1.0 | 71.6 | 0.000 | 0.000 | 10 | 10 | 0 |
| 125 | 25 | 0.2 | 4.2 | 14.6 | 0.346 | 4.0 | 16.8 | 0.000 | 0.000 | 10 | 8 | 6 |
| 125 | 25 | 0.4 | 2.1 | 36.9 | 0.213 | 2.1 | 41.1 | 0.000 | 0.000 | 10 | 10 | 9 |
| 125 | 25 | 0.6 | 1.5 | 57.5 | 0.008 | 1.5 | 57.5 | 0.000 | 0.000 | 10 | 10 | 0 |
| 125 | 25 | 0.8 | 1.0 | 72.7 | 0.005 | 1.0 | 72.7 | 0.000 | 0.000 | 10 | 10 | 0 |
| 125 | 62 | 0.2 | 5.0 | 15.1 | 5.798 | 4.8 | 16.7 | 0.000 | 0.000 | 10 | 8 | 5 |
| 125 | 62 | 0.4 | 3.0 | 30.7 | 1.875 | 2.8 | 34.4 | 0.000 | 0.000 | 10 | 8 | 4 |
| 125 | 62 | 0.6 | 2.0 | 47.0 | 0.019 | 2.0 | 47.7 | 0.000 | 0.000 | 10 | 10 | 5 |
| 125 | 62 | 0.8 | 1.0 | 76.2 | 0.007 | 1.0 | 76.2 | 0.000 | 0.000 | 10 | 10 | 0 |
| 125 | 125 | 0.2 | 5.6 | 15.2 | 69.342 | 5.0 | 17.5 | 0.001 | 0.000 | 10 | 4 | 3 |
| 125 | 125 | 0.4 | 3.0 | 30.9 | 7.642 | 3.0 | 33.7 | 0.000 | 0.000 | 10 | 10 | 9 |
| 125 | 125 | 0.6 | 2.0 | 48.0 | 0.120 | 2.0 | 50.0 | 0.000 | 0.000 | 10 | 10 | 8 |
| 125 | 125 | 0.8 | 1.0 | 76.5 | 0.015 | 1.0 | 76.5 | 0.001 | 0.000 | 10 | 10 | 0 |
| 250 | 25 | 0.2 | 4.0 | 28.2 | 0.705 | 3.9 | 30.8 | 0.000 | 0.000 | 10 | 9 | 3 |
| 250 | 25 | 0.4 | 2.0 | 70.6 | 0.281 | 2.0 | 81.2 | 0.000 | 0.000 | 10 | 10 | 10 |
| 250 | 25 | 0.6 | 1.2 | 129.8 | 0.002 | 1.2 | 129.8 | 0.000 | 0.000 | 10 | 10 | 0 |
| 250 | 25 | 0.8 | 1.0 | 141.4 | 0.006 | 1.0 | 141.4 | 0.000 | 0.000 | 10 | 10 | 0 |
| 250 | 50 | 0.2 | 4.0 | 28.5 | 10.367 | 4.0 | 33.9 | 0.002 | 0.000 | 10 | 10 | 10 |
| 250 | 50 | 0.4 | 2.1 | 70.7 | 2.519 | 2.1 | 81.8 | 0.000 | 0.000 | 10 | 10 | 9 |

| n | r | α | |S*| | |N_v(S*)| | CPU | |S*| | |N_v(S*)| | CPU | CPU | #Opt | #E1 | #E2 |
|---|---|---|---|---|---|---|---|---|---|---|---|---|
| 250 | 50 | 0.6 | 1.5 | 115.7 | 0.008 | 1.5 | 115.7 | 0.000 | 0.000 | 10 | 10 | 0 |
| 250 | 50 | 0.8 | 1.0 | 143.8 | 0.006 | 1.0 | 143.8 | 0.000 | 0.000 | 10 | 10 | 0 |
| 250 | 125 | 0.2 | 4.2 | 29.7 | 727.591 | 4.1 | 36.0 | 0.002 | 0.001 | 9 | 9 | 8 |
| 250 | 125 | 0.4 | 2.4 | 68.6 | 33.135 | 2.3 | 79.0 | 0.000 | 0.000 | 10 | 9 | 7 |
| 250 | 125 | 0.6 | 1.8 | 100.3 | 0.035 | 1.8 | 100.6 | 0.001 | 0.000 | 10 | 10 | 2 |
| 250 | 125 | 0.8 | 1.0 | 145.7 | 0.016 | 1.0 | 145.7 | 0.000 | 0.000 | 10 | 10 | 0 |
| 250 | 250 | 0.2 | 4.1 | 29.9 | 262.017 | 4.3 | 36.2 | 0.001 | 0.001 | 0 | 8 | 8 |
| 250 | 250 | 0.4 | 2.7 | 64.4 | 310.897 | 2.7 | 68.7 | 0.002 | 0.001 | 8 | 10 | 7 |
| 250 | 250 | 0.6 | 1.8 | 101.5 | 0.149 | 1.8 | 103.4 | 0.001 | 0.001 | 10 | 10 | 6 |
| 250 | 250 | 0.8 | 1.0 | 147.7 | 0.128 | 1.0 | 147.7 | 0.001 | 0.000 | 10 | 10 | 0 |
| 500 | 50 | 0.2 | 3.8 | 58.0 | 46.154 | 3.4 | 76.5 | 0.001 | 0.000 | 10 | 6 | 4 |
| 500 | 50 | 0.4 | 2.0 | 140.7 | 3.880 | 2.0 | 158.3 | 0.000 | 0.000 | 10 | 10 | 10 |
| 500 | 50 | 0.6 | 1.0 | 276.3 | 0.019 | 1.0 | 276.3 | 0.000 | 0.000 | 10 | 10 | 0 |
| 500 | 50 | 0.8 | 1.0 | 276.3 | 0.019 | 1.0 | 276.3 | 0.000 | 0.000 | 10 | 10 | 0 |
| 500 | 100 | 0.2 | 3.6 | 59.1 | 318.040 | 3.8 | 64.7 | 0.002 | 0.001 | 0 | 8 | 8 |
| 500 | 100 | 0.4 | 2.0 | 136.8 | 149.210 | 2.0 | 160.1 | 0.001 | 0.000 | 10 | 10 | 10 |
| 500 | 100 | 0.6 | 1.1 | 267.6 | 0.030 | 1.1 | 267.6 | 0.000 | 0.000 | 10 | 10 | 0 |
| 500 | 100 | 0.8 | 1.0 | 278.4 | 0.028 | 1.0 | 278.4 | 0.000 | 0.000 | 10 | 10 | 0 |
| 500 | 250 | 0.2 | 3.6 | 61.9 | 100.025 | 4.0 | 61.8 | 0.003 | 0.002 | 0 | 6 | 6 |
| 500 | 250 | 0.4 | 2.0 | 142.5 | 100.023 | 2.0 | 166.1 | 0.003 | 0.001 | 0 | 10 | 10 |
| 500 | 250 | 0.6 | 1.3 | 248.1 | 0.055 | 1.3 | 248.1 | 0.001 | 0.000 | 10 | 10 | 0 |
| 500 | 250 | 0.8 | 1.0 | 282.0 | 0.053 | 1.0 | 282.0 | 0.002 | 0.000 | 10 | 10 | 0 |
| 500 | 500 | 0.2 | 3.4 | 65.2 | 120.054 | 4.0 | 66.6 | 0.007 | 0.003 | 0 | 4 | 4 |
| 500 | 500 | 0.4 | 2.0 | 144.8 | 120.050 | 2.0 | 168.3 | 0.003 | 0.002 | 0 | 10 | 10 |
| 500 | 500 | 0.6 | 1.4 | 238.7 | 4.306 | 1.4 | 238.7 | 0.004 | 0.001 | 10 | 10 | 0 |
| 500 | 500 | 0.8 | 1.0 | 284.2 | 3.589 | 1.0 | 284.2 | 0.001 | 0.000 | 10 | 10 | 0 |
| **Average** | | | **2.2** | **105.9** | **49.981** | **2.2** | **110.1** | **0.001** | **0.000** | **8.7** | **9.3** | **4.0** |

Table 2.2: Comparison of MIP, Heuristic and Accelerated Heuristic for $v = 1$

| | | | MIP | | | H | | | AH | | | |
|---|---|---|---|---|---|---|---|---|---|---|---|---|
| n | r | α | |S*| | |N_v(S*)| | CPU | |S*| | |N_v(S*)| | CPU | CPU | #Opt | #E1 | #E2 |
| 125 | 12 | 0.2 | 3.4 | 29.8 | 0.171 | 3.4 | 33.6 | 0.000 | 0.000 | 10 | 10 | 5 |
| 125 | 12 | 0.4 | 2.0 | 65.1 | 0.195 | 2.0 | 70.0 | 0.000 | 0.000 | 10 | 10 | 8 |
| 125 | 12 | 0.6 | 1.1 | 120.0 | 0.004 | 1.1 | 120.0 | 0.000 | 0.000 | 10 | 10 | 0 |
| 125 | 12 | 0.8 | 1.0 | 125.0 | 0.007 | 1.0 | 125.0 | 0.000 | 0.000 | 10 | 10 | 0 |
| 125 | 25 | 0.2 | 4.1 | 25.6 | 0.699 | 3.8 | 31.1 | 0.000 | 0.000 | 10 | 7 | 4 |
| 125 | 25 | 0.4 | 2.0 | 64.2 | 0.295 | 2.0 | 73.8 | 0.000 | 0.000 | 10 | 10 | 9 |
| 125 | 25 | 0.6 | 1.5 | 100.7 | 0.010 | 1.5 | 100.7 | 0.000 | 0.000 | 10 | 10 | 0 |
| 125 | 25 | 0.8 | 1.0 | 125.0 | 0.005 | 1.0 | 125.0 | 0.000 | 0.000 | 10 | 10 | 0 |
| 125 | 62 | 0.2 | 4.1 | 25.3 | 20.880 | 4.0 | 30.3 | 0.000 | 0.000 | 10 | 9 | 9 |
| 125 | 62 | 0.4 | 2.1 | 61.7 | 4.091 | 2.1 | 71.4 | 0.000 | 0.000 | 10 | 10 | 9 |
| 125 | 62 | 0.6 | 1.4 | 105.4 | 0.011 | 1.4 | 105.5 | 0.000 | 0.000 | 10 | 10 | 1 |
| 125 | 62 | 0.8 | 1.0 | 125.0 | 0.010 | 1.0 | 125.0 | 0.000 | 0.000 | 10 | 10 | 0 |
| 125 | 125 | 0.2 | 4.5 | 25.2 | 453.025 | 4.0 | 32.4 | 0.001 | 0.000 | 10 | 5 | 5 |
| 125 | 125 | 0.4 | 2.1 | 62.3 | 18.843 | 2.1 | 74.2 | 0.001 | 0.000 | 10 | 10 | 9 |
| 125 | 125 | 0.6 | 1.8 | 86.3 | 0.020 | 1.8 | 86.8 | 0.000 | 0.000 | 10 | 10 | 3 |
| 125 | 125 | 0.8 | 1.0 | 125.0 | 0.015 | 1.0 | 125.0 | 0.001 | 0.000 | 10 | 10 | 0 |
| 250 | 25 | 0.2 | 3.3 | 56.6 | 1.730 | 3.3 | 70.5 | 0.000 | 0.000 | 10 | 10 | 9 |
| 250 | 25 | 0.4 | 2.0 | 127.0 | 0.600 | 2.0 | 141.0 | 0.000 | 0.000 | 10 | 10 | 10 |
| 250 | 25 | 0.6 | 1.1 | 240.2 | 0.009 | 1.1 | 240.2 | 0.000 | 0.000 | 10 | 10 | 0 |

| 250 | 25 | 0.8 | 1.0 | 250.0 | 0.011 | 1.0 | 250.0 | 0.000 | 0.000 | 10 | 10 | 0 |
|---|---|---|---|---|---|---|---|---|---|---|---|---|
| 250 | 50 | 0.2 | 3.9 | 51.4 | 30.994 | 3.7 | 61.3 | 0.000 | 0.000 | 10 | 8 | 8 |
| 250 | 50 | 0.4 | 2.0 | 126.3 | 3.518 | 2.0 | 143.4 | 0.000 | 0.000 | 10 | 10 | 10 |
| 250 | 50 | 0.6 | 1.0 | 250.0 | 0.016 | 1.0 | 250.0 | 0.000 | 0.000 | 10 | 10 | 0 |
| 250 | 50 | 0.8 | 1.0 | 250.0 | 0.016 | 1.0 | 250.0 | 0.000 | 0.000 | 10 | 10 | 0 |
| 250 | 125 | 0.2 | 4.0 | 50.9 | 2279.830 | 4.0 | 56.1 | 0.003 | 0.002 | 1 | 10 | 10 |
| 250 | 125 | 0.4 | 2.0 | 123.5 | 182.474 | 2.0 | 145.6 | 0.000 | 0.000 | 10 | 10 | 10 |
| 250 | 125 | 0.6 | 1.0 | 250.0 | 0.033 | 1.0 | 250.0 | 0.000 | 0.000 | 10 | 10 | 0 |
| 250 | 125 | 0.8 | 1.0 | 250.0 | 0.031 | 1.0 | 250.0 | 0.000 | 0.000 | 10 | 10 | 0 |
| 250 | 250 | 0.2 | 3.7 | 53.0 | 248.032 | 4.0 | 57.8 | 0.002 | 0.001 | 0 | 7 | 7 |
| 250 | 250 | 0.4 | 2.0 | 124.5 | 2166.380 | 2.0 | 145.6 | 0.001 | 0.000 | 0 | 10 | 10 |
| 250 | 250 | 0.6 | 1.0 | 250.0 | 0.064 | 1.0 | 250.0 | 0.000 | 0.000 | 10 | 10 | 0 |
| 250 | 250 | 0.8 | 1.0 | 250.0 | 0.054 | 1.0 | 250.0 | 0.001 | 0.000 | 10 | 10 | 0 |
| 500 | 50 | 0.2 | 3.0 | 119.0 | 262.746 | 3.0 | 153.6 | 0.000 | 0.000 | 10 | 10 | 10 |
| 500 | 50 | 0.4 | 2.0 | 249.2 | 46.086 | 2.0 | 274.6 | 0.001 | 0.000 | 10 | 10 | 10 |
| 500 | 50 | 0.6 | 1.0 | 500.0 | 0.029 | 1.0 | 500.0 | 0.001 | 0.001 | 10 | 10 | 0 |
| 500 | 50 | 0.8 | 1.0 | 500.0 | 0.029 | 1.0 | 500.0 | 0.000 | 0.000 | 10 | 10 | 0 |
| 500 | 100 | 0.2 | 3.0 | 118.3 | 298.100 | 3.0 | 155.9 | 0.003 | 0.002 | 0 | 10 | 10 |
| 500 | 100 | 0.4 | 2.0 | 249.6 | 1203.750 | 2.0 | 276.1 | 0.001 | 0.000 | 10 | 10 | 10 |
| 500 | 100 | 0.6 | 1.0 | 500.0 | 0.040 | 1.0 | 500.0 | 0.000 | 0.000 | 10 | 10 | 0 |
| 500 | 100 | 0.8 | 1.0 | 500.0 | 0.040 | 1.0 | 500.0 | 0.000 | 0.000 | 10 | 10 | 0 |
| 500 | 250 | 0.2 | 3.0 | 121.7 | 100.034 | 3.2 | 151.1 | 0.003 | 0.001 | 0 | 8 | 8 |
| 500 | 250 | 0.4 | 2.0 | 248.4 | 100.051 | 2.0 | 278.9 | 0.000 | 0.000 | 0 | 10 | 10 |
| 500 | 250 | 0.6 | 1.0 | 500.0 | 0.206 | 1.0 | 500.0 | 0.004 | 0.001 | 10 | 10 | 0 |
| 500 | 250 | 0.8 | 1.0 | 500.0 | 0.233 | 1.0 | 500.0 | 0.004 | 0.000 | 10 | 10 | 0 |
| 500 | 500 | 0.2 | 2.8 | 162.9 | 116.086 | 3.5 | 132.7 | 0.008 | 0.005 | 0 | 4 | 4 |
| 500 | 500 | 0.4 | 1.8 | 294.1 | 120.250 | 2.0 | 280.4 | 0.008 | 0.004 | 0 | 8 | 8 |
| 500 | 500 | 0.6 | 1.0 | 500.0 | 0.284 | 1.0 | 500.0 | 0.006 | 0.000 | 10 | 10 | 0 |
| 500 | 500 | 0.8 | 1.0 | 500.0 | 0.296 | 1.0 | 500.0 | 0.007 | 0.000 | 10 | 10 | 0 |
| **Average** | | | **1.9** | **198.7** | **159.590** | **1.9** | **205.1** | **0.001** | **0.000** | **8.4** | **9.5** | **4.3** |

Table 2.3: Comparison of MIP, Heuristic and Accelerated Heuristic for $v = 01$

Tables 3.1, 3.2 and 3.3 compare the results of the MIP solver (CPLEX), the constructive heuristic (H) and its accelerated version (AH) for value $v = 0$, 1 and 01 respectively, on large-sized instances with $n = 2000$ and $r = 50$. Each line corresponds to the average of 4 randomly generated instances for a fixed value of $n$, $r$. For $v = 0$ (see Table 3.1), 94% of the instances are solved optimally except for 4 instances over 72 (i.e. only one instance is not solved optimally due to the memory limitation "Out of Memory" error and the other 3 instances are not solved optimally in the time limit of one hour). Table 3.1 shows also that the constructive heuristic reaches an optimal value for 71% of the instances and provides solutions that dominate those provided by CPLEX in 64% of the cases. For $v = 1$ (see Table 3.2), all 72 instances are solved optimally. Table 3.2 shows that the constructive heuristic reaches optimal value for 96% of the instances and provides solutions that dominate those provided by CPLEX in 33% of the instances. For $v = 01$ (see Table 3.3), 79% of the instances are solved optimally except for 15 instances over 72 (i.e. 3 instances are not solved optimally in the time limit of one hour and the other 12 instances are not solved optimally due to the memory limitation "Out of Memory" error). The constructive heuristic reaches an optimal value for 76% of the instances and provides solutions that dominate those provided by CPLEX in 56% of the instances.

|  | MIP | | | H | | | AH | | | |
| --- | --- | --- | --- | --- | --- | --- | --- | --- | --- | --- |
| Instance | $|S^*|$ | $|N_v(S^*)|$ | CPU | $|S^*|$ | $|N_v(S^*)|$ | CPU | CPU | #Opt | #E1 | #E2 |
| 1 | 14.50 | 1274.50 | 197.730 | 14.50 | 1283.75 | 0.007 | 0.004 | 3 | 4 | 4 |
| 2 | 16.00 | 1317.50 | 151.897 | 14.50 | 1326.50 | 0.008 | 0.005 | 4 | 1 | 1 |
| 3 | 14.75 | 1229.50 | 1123.480 | 13.75 | 1235.75 | 0.005 | 0.003 | 3 | 1 | 1 |
| 4 | 16.50 | 1169.25 | 300.759 | 16.00 | 1177.50 | 0.010 | 0.003 | 4 | 2 | 2 |
| 5 | 16.00 | 1228.00 | 317.072 | 16.00 | 1238.75 | 0.005 | 0.005 | 4 | 4 | 4 |
| 6 | 16.50 | 1241.00 | 489.668 | 16.50 | 1256.00 | 0.008 | 0.000 | 4 | 4 | 4 |
| 7 | 18.50 | 1313.00 | 169.470 | 18.50 | 1320.75 | 0.008 | 0.005 | 4 | 4 | 3 |
| 8 | 16.50 | 1200.00 | 384.925 | 16.50 | 1207.75 | 0.007 | 0.003 | 4 | 4 | 4 |
| 9 | 15.00 | 1244.50 | 1055.460 | 15.00 | 1260.25 | 0.010 | 0.003 | 3 | 4 | 3 |
| 10 | 15.00 | 1256.25 | 1099.630 | 14.25 | 1264.75 | 0.005 | 0.000 | 3 | 1 | 1 |
| 11 | 16.50 | 1223.50 | 212.398 | 16.50 | 1235.50 | 0.008 | 0.008 | 4 | 4 | 4 |
| 12 | 16.00 | 1220.00 | 435.857 | 15.00 | 1233.75 | 0.008 | 0.003 | 4 | 1 | 1 |
| 13 | 16.25 | 1181.00 | 586.565 | 16.25 | 1185.75 | 0.008 | 0.003 | 4 | 4 | 2 |
| 14 | 15.25 | 1330.00 | 243.923 | 14.75 | 1335.25 | 0.005 | 0.000 | 4 | 2 | 1 |
| 15 | 15.25 | 1255.25 | 604.018 | 15.00 | 1267.75 | 0.008 | 0.003 | 4 | 3 | 3 |
| 16 | 17.00 | 1270.75 | 278.472 | 15.50 | 1290.50 | 0.008 | 0.000 | 4 | 2 | 2 |
| 17 | 15.75 | 1296.00 | 253.367 | 16.00 | 1312.75 | 0.006 | 0.003 | 4 | 3 | 3 |
| 18 | 17.50 | 1265.75 | 157.707 | 17.25 | 1273.75 | 0.008 | 0.003 | 4 | 3 | 3 |
| **Avg** | **16.04** | **1250.88** | **447.911** | **15.65** | **1261.49** | **0.007** | **0.003** | **3.78** | **2.83** | **2.56** |

Table 3.1: Comparison of MIP, Heuristic and Accelerated Heuristic for $n$=2000, $r$=50 and $v$=0

|  | MIP | | | H | | | AH | | | |
| --- | --- | --- | --- | --- | --- | --- | --- | --- | --- | --- |
| Instance | $|S^*|$ | $|N_v(S^*)|$ | CPU | $|S^*|$ | $|N_v(S^*)|$ | CPU | CPU | #Opt | #E1 | #E2 |
| 1 | 2.00 | 130.50 | 0.974 | 2.00 | 130.50 | 0.001 | 0.001 | 4 | 4 | 0 |
| 2 | 2.25 | 132.25 | 1.576 | 2.00 | 137.25 | 0.000 | 0.000 | 4 | 3 | 1 |
| 3 | 1.50 | 160.00 | 0.155 | 1.50 | 160.00 | 0.003 | 0.000 | 4 | 4 | 0 |
| 4 | 2.00 | 134.75 | 1.181 | 2.00 | 137.00 | 0.000 | 0.000 | 4 | 4 | 2 |
| 5 | 2.00 | 132.75 | 1.184 | 2.00 | 135.00 | 0.000 | 0.000 | 4 | 4 | 2 |
| 6 | 2.25 | 133.75 | 1.628 | 2.25 | 134.00 | 0.000 | 0.000 | 4 | 4 | 1 |
| 7 | 2.50 | 132.75 | 6.286 | 2.50 | 138.00 | 0.003 | 0.000 | 4 | 4 | 2 |
| 8 | 2.00 | 132.50 | 1.310 | 2.00 | 138.00 | 0.001 | 0.000 | 4 | 4 | 2 |
| 9 | 2.00 | 133.75 | 0.675 | 2.00 | 134.50 | 0.000 | 0.000 | 4 | 4 | 1 |
| 10 | 1.75 | 132.00 | 0.510 | 1.75 | 137.00 | 0.003 | 0.000 | 4 | 4 | 1 |
| 11 | 2.25 | 130.75 | 0.786 | 2.00 | 137.00 | 0.000 | 0.000 | 4 | 3 | 1 |
| 12 | 2.00 | 131.75 | 0.582 | 2.00 | 135.75 | 0.005 | 0.000 | 4 | 4 | 2 |
| 13 | 1.75 | 135.00 | 0.190 | 1.75 | 136.75 | 0.003 | 0.000 | 4 | 4 | 1 |
| 14 | 2.00 | 133.75 | 1.135 | 2.00 | 135.75 | 0.000 | 0.000 | 4 | 4 | 1 |
| 15 | 2.00 | 131.50 | 1.723 | 2.00 | 136.50 | 0.000 | 0.000 | 4 | 4 | 2 |
| 16 | 2.25 | 134.50 | 0.403 | 2.25 | 138.50 | 0.003 | 0.000 | 4 | 4 | 2 |
| 17 | 2.75 | 130.25 | 2.208 | 2.50 | 138.00 | 0.002 | 0.000 | 4 | 3 | 1 |

| | | | | | | | | | |
|---|---|---|---|---|---|---|---|---|---|
| 18 | 2.50 | 132.00 | 2.997 | 2.50 | 138.25 | 0.000 | 0.000 | 4 | 4 | 2 |
| Avg | 2.10 | 134.14 | 1.417 | 2.06 | 137.65 | 0.001 | 0.000 | 4.00 | 3.83 | 1.33 |

Table 3.2: Comparison of MIP, Heuristic and Accelerated Heuristic for $n=2000$, $r=50$ and $v=1$

| | MIP | | | H | | | AH | | | |
|---|---|---|---|---|---|---|---|---|---|---|
| Instance | $|S^*|$ | $|N_v(S^*)|$ | CPU | $|S^*|$ | $|N_v(S^*)|$ | CPU | CPU | #Opt | #E1 | #E2 |
| 1 | 12.00 | 1371.75 | 380.550 | 12.00 | 1384.50 | 0.007 | 0.005 | 3 | 4 | 4 |
| 2 | 13.00 | 1427.25 | 514.354 | 12.00 | 1429.75 | 0.005 | 0.003 | 4 | 2 | 1 |
| 3 | 11.75 | 1396.50 | 511.151 | 11.25 | 1400.00 | 0.003 | 0.005 | 3 | 2 | 1 |
| 4 | 14.00 | 1283.00 | 652.712 | 13.50 | 1294.00 | 0.008 | 0.000 | 3 | 3 | 2 |
| 5 | 13.25 | 1330.25 | 309.424 | 13.25 | 1348.50 | 0.005 | 0.003 | 3 | 4 | 3 |
| 6 | 13.75 | 1345.25 | 222.256 | 13.75 | 1366.50 | 0.005 | 0.003 | 3 | 2 | 2 |
| 7 | 15.00 | 1417.00 | 936.245 | 15.00 | 1431.25 | 0.006 | 0.003 | 4 | 4 | 2 |
| 8 | 13.75 | 1320.25 | 257.064 | 13.75 | 1323.25 | 0.005 | 0.005 | 3 | 4 | 2 |
| 9 | 12.75 | 1351.25 | 1037.430 | 12.75 | 1363.50 | 0.003 | 0.003 | 2 | 4 | 4 |
| 10 | 12.00 | 1361.75 | 555.008 | 11.75 | 1368.00 | 0.008 | 0.000 | 3 | 3 | 2 |
| 11 | 14.00 | 1320.75 | 1186.710 | 14.00 | 1336.75 | 0.008 | 0.003 | 3 | 4 | 3 |
| 12 | 13.25 | 1326.25 | 307.274 | 12.50 | 1352.00 | 0.007 | 0.000 | 3 | 1 | 1 |
| 13 | 13.75 | 1284.50 | 714.217 | 13.50 | 1314.00 | 0.003 | 0.005 | 4 | 3 | 2 |
| 14 | 12.75 | 1434.25 | 137.308 | 12.25 | 1437.25 | 0.005 | 0.003 | 3 | 2 | 1 |
| 15 | 12.50 | 1365.75 | 782.128 | 12.25 | 1380.25 | 0.005 | 0.005 | 3 | 3 | 3 |
| 16 | 14.00 | 1385.50 | 222.542 | 13.25 | 1389.25 | 0.007 | 0.006 | 3 | 2 | 1 |
| 17 | 13.50 | 1407.75 | 760.556 | 13.50 | 1417.25 | 0.005 | 0.004 | 4 | 4 | 3 |
| 18 | 14.50 | 1368.25 | 9675.590 | 14.50 | 1389.25 | 0.005 | 0.008 | 3 | 4 | 3 |
| Avg | 13.31 | 1360.96 | 1064.584 | 13.04 | 1373.63 | 0.005 | 0.003 | 3.17 | 3.06 | 2.22 |

Table 3.3: Comparison of MIP, Heuristic and Accelerated Heuristic for $n=2000$, $r=50$ and $v=01$

Table 4 compares the CPU time consumed by the constructive heuristic (H) and its accelerated version (AH) for value $v = 0$, 1 and 01, on large-sized instances with $n = 2000$ and $r \in \{100, 200\}$. Each line corresponds to the average of 4 randomly generated instances for a fixed value of $r$. For $v = 0$ and $v = 1$, the CPU time of the accelerated heuristic is half of the CPU time consumed by the constructive heuristic on average.

| | | $v = 0$ | | $v = 1$ | | $v = 01$ | |
|---|---|---|---|---|---|---|---|
| $r$ | $\alpha$ | H | AH | H | AH | H | AH |
| 100 | 0.2 | 0.004 | 0.003 | 0.004 | 0.002 | 0.004 | 0.003 |
| 100 | 0.4 | 0.003 | 0.002 | 0.003 | 0.002 | 0.003 | 0.003 |
| 100 | 0.6 | 0.002 | 0.001 | 0.002 | 0.001 | 0.002 | 0.002 |
| 100 | 0.8 | 0.002 | 0.001 | 0.002 | 0.001 | 0.002 | 0.002 |
| 200 | 0.2 | 0.007 | 0.005 | 0.007 | 0.005 | 0.008 | 0.006 |
| 200 | 0.4 | 0.006 | 0.004 | 0.006 | 0.004 | 0.006 | 0.005 |
| 200 | 0.6 | 0.004 | 0.002 | 0.003 | 0.002 | 0.004 | 0.003 |
| 200 | 0.8 | 0.004 | 0.002 | 0.004 | 0.002 | 0.003 | 0.003 |

| Average | 0.004 | 0.002 | 0.004 | 0.002 | 0.004 | 0.003 |

Table 4: CPU time comparison of Heuristic and Accelerated Heuristic for *n*=2000

## 8. Conclusions

In addition to applications of our algorithms to clustering, the exploitation of consistency by the approaches described above open a number of opportunities for applying our algorithms to generate reference sets for evolutionary metaheuristics. Questions related to such opportunities include the following, whose answers will depend in part on the evolutionary metaheuristics employed.

1. Which values of $v \in \{0, 1, 01\}$ prove most useful for particular classes of optimization problems?
2. What values for the bounds $L_v^1$ and $L_v^2$ for Problems $G_v^1$ and $G_v^2$ are best (e.g., as a function of *n*)?
3. Which approximation algorithm should be used to generate the set *R* (as in Step 1 of the SD&C Algorithm) and what size should be selected for *R*?
4. Should the approximation algorithm be run only for a small number of iterations to generate new instances of *R* after the first, and what portion of a previous *R* should be saved to merge with a new *R*?
5. What type of diversification approach should be used as a foundation for generating instances of *R* after the first instance?
6. Can better sets *S* be generated by using tie-breaking that favors higher *f(x)* values, or more generally, by using the parameter α as suggested in Section 5?
7. What range of different sets *S* may be generated to give useful variation in the consistent variables identified and to identify strongly determined variables by higher order frequencies?
8. Do advantages result by using an adaptive version of Algorithms 1 and 2 as indicated in Section 6?

Metaheuristic strategies found in Palubekis (2004), Glover (2016), Lai, et al. (2016), Wang, et al. (2012, 2017, 2018), Glover and Hao (2018) are relevant to exploring the issues raised by the preceding questions. However, the new considerations introduced in this paper invite empirical studies of forms not previously conducted.

## Appendix 1.

We begin by describing Algorithms 1 and 2 for problems $G_v^1$ and $G_v^2$ using set notation in place of the vector notation used in Section 4. Then we give a detailed version of Algorithm 1 that incorporates aspects of both the set notation and vector notation and is organized for greater efficiency.

**Outline for Algorithm 1 for Problem $G_v^1$ (Maximize |S|: for $x^1 \in S$ and $L_v^1 \leq |N_v(S)|$)**
Initialize k = 1 and S = $\{x^1\}$.
For successive iterations k = 1, 2, …, while $R^o \neq \emptyset$
    Select a vector x ∈ $R^o$ that maximizes $|N_v(S \cup \{x\})|$ for x ∈ $R_o$, and denote the chosen
    vector by x*. Hence
    x* = arg max($|N_v(S \cup \{x\})|$: x ∈ $R_o$)
    Let S* = S∪{x*}, $N_v^* = N_v(S^*)$ and $n_v^* = |N_v^*|$
    $N_v(S \cup \{x\}) = N_v(S) \cap N_v(x)$
    Save: $N_v = N_v(S)$ and find x* to maximize $|N_v \cap N_v(x)|$
Continue next iteration for k := k + 1)

This gives rise to the following expanded description.

**Algorithm 1:**
Choose $x^1 \in R$; $R^o = R \setminus \{x^1\}$; k = 1; $N_v = N_v(x^1)$; $n_v = |N_v|$; Done = False
While Done = False and $R^o \neq \emptyset$ do
    For each x ∈ $R^o$ do
        $N_v^o = N_v \cap N_v(x)$

$n_v = |N_v^o|$
                If $n_v > n_v^*$ then
                        $n_v^* = n_v$
                        $N_v^* = N_v^o$
                        $x^* = x$
                        $x_o^* = f(x^*)$
                Endif
        Endfor
        **Execute the Update/Terminate Routine**
        If $n_v^* \geq L_v^1$
                $S^* = S \cup \{x^*\}$
                $k := k + 1$ % $k = |S^*|$
                $x^k = x^*$
                $N_v = N_v^* \ (= N_v(S^*))$
                $R^o = R^o \setminus \{x^*\}$
        Else
                Done = True
        Endif
Endwhile

As indicated in Section 4, the record $x_o^* = f(x^*)$ is for the option of breaking ties when $n_v = n_v^*$ to favor solutions with larger objective values $f(x)$. In this case the instruction "If $n_v > n_v^*$" can be enlarged to become "If $n_v > n_v^*$ or if $n_v = n_v^*$ and $f(x) > x_o^*$."

**Outline for Algorithm 2 for Problem $G_v^2$ (Maximize $|N_v(S)|$: for $x^1 \in S$ and $L_v^2 \leq |S|$.)**
Initialize $k = 1$ and $S = \{x^1\}$.
For successive iterations $k = 1, 2, \ldots$, while $R^o \neq \emptyset$
        Select a vector $x \in R^o$ that maximizes $|z|_v$ for $z = z^k \cap_o x$, and denote the chosen vector
         by $x^*$. Hence
        $x^* = \arg \max(|z|_v : z = z^k \cap_o x, x \in R^o)$
        $z^* = z^k \cap_o x^*$ ($z^*$ is the vector analog of $S \cup \{x^*\}$.)
        $S := S \cup \{x^*\}$, $z^{k+1} = z^*$ and $R^o := R^o \setminus \{x^*\}$.
        If $L_v^2 = |S|$ ( $= k + 1$) or $R^o = \emptyset$. stop with the current S.
Continue next iteration for $k := k + 1$

This gives rise to the following expanded description.

**Algorithm 2.**
Choose $x^1 \in R$; $R^o = R \setminus \{x^1\}$; $k = 1$; $N_v = N_v(x)$; $n_v = |N_v|$; Done = False
While Done = False and $R^o \neq \emptyset$
        For each $x \in R^o$
                $N_v^o = N_v \cap N_v(x)$
                $n_v = |N_v^o|$
                If $n_v > n_v^*$
                        $n_v^* = n_v$

$$N_v^* = N_v^o$$
$$x^* = x$$
$$x_o^* = f(x^*)$$
     Endif
    Endfor
    **Execute the Update/Terminate Routine for Problem $G_v^2$**
    If $n_v^* \geq L_v^1$
      $S^* = S \cup \{x^*\}$
      $k := k + 1$ % $k = |S^*|$
      $x^k = x^*$
      $N_v = N_v^*$ $(= N_v(S^*))$
      $R^o = R^o \backslash \{x^*\}$
    Else
      Done = True
    Endif
  Endwhile

Now we give the detailed description of Algorithm 1 organized for efficiency.

## Detailed Version of Algorithm 1

*Given*:
R and choice of v = 0, 1, or 01 and the value $L_v^1$, where $L_v^1$ cannot exceed the value $n_v$ in the following Initialization.

*Initialization:*
Select $x^1$ from R.
$R^o = R \backslash \{x^1\}$, $z^1 = x^1$, $N_{01} = N$, $N_0 = \{j \in N: x_j^1 = 0\}$, $N_1 = \{j \in N: x_j^1 = 1\}$, $N^\# = \emptyset$.
$n_v = |N_v|$ for v = 0, 1 and 01
$k = 1$ ($k = |S|$)
$x^* = x^1$, $N_{01}^* = N_{01}$, $N_0^* = N_0$, $N_1^* = N_1$ (saves the best current candidate solution)
$n_v^* = 0$ (This assures that the first non-empty set $N_v$ will qualify as a candidate to add to the current S.)

While $R^o \neq \emptyset$ do
  While some solution in $R^o$ remains unexamined do
    Select $x = x(h)$ from $R^o$
    $N_0 Temp = N_0$, $N_1 Temp = N_1$,
    $N_{01} Temp = N_{01}$, $N^\# Temp = N^\#$
    $n_v Temp = n_v$ for v = 0, 1 and 01
    For each $j \in N_0$: (This need only be executed for v = 0 or 01) do
      If $x_j = 0$ then
        $z_j = 0$
      Else (if $x_j = 1$)
        $z_j = \#$
        $N_0 Temp := N_0 Temp \backslash \{j\}$ and $N_{01} Temp := N_{01} Temp \backslash \{j\}$
        $n_o Temp := n_0 Temp - 1$ and $n_{01} Temp := Temp n_{01} - 1$

$N^{\#}$Temp := $N^{\#}$Temp$\cup\{j\}$; $n^{\#}$Temp := $n^{\#}$Temp + 1
If $n_v$Temp < $L_v^1$ when v = 0 or 01 then
   Break to **Update for Problem $G_v^1$**
            Endif
      Endfor
      For each j ∈ $N_1$: (This need only be executed for v = 1 or 01) do
            If $x_j$ = 1 then
                  $z_j$ = 1
            Else (if $x_j$ = 0)
                  $z_j$ = #
                  $N_1$Temp := $N_1$Temp $\setminus \{j\}$ and $N_{01}$Temp := $N_{01}$Temp$\setminus\{j\}$
                  $n_1$Temp := $n_1$Temp − 1 ; $n_{01}$Temp := $n_{01}$Temp − 1
                  $N^{\#}$Temp := $N^{\#}$Temp$\cup\{j\}$; $n^{\#}$Temp := $n^{\#}$Temp + 1
                  If $n_v$Temp < $L_v^1$ when v = 1 or 01 then
                        Break to **Update for Problem $G_v^1$**
            Endif
      Endfor
      % The assignments $z_j$ = # and the update of $N^{\#}$Temp above are not strictly
       necessary. They can always be identified by reference to the other information
       stored, as indicated subsequently.
      % Next evaluates x by reference to $n_v$Temp.

      **Update For Problem $G_v^1$.**
      If $n_v$Temp < $L_v^1$ then
            % x is not feasible for Problem $G_v^1$ and will never be feasible on later
             iterations of the Inner While loop.
            $R^o$ := $R^o\setminus\{x\}$
      Else
            % x is feasible and hence check whether it is best among those solutions in
             $R^o$ examined so far
            If $n_v$Temp > $n_v^*$ then
                  $h^*$ = h for $x^*$ = x(h) (gives a simple way to recover $x^*$)
                  % The new best solution $x^*$ creates the following sets with
                   their indicated cardinalities $n_0^*$, $n_1^*$ and $n_{01}^*$.
                  $N_0^*$ = $N_0$Temp, $N_1^*$ = $N_1$Temp and $N_{01}^*$ = $N_{01}$Temp
                  $n_0^*$ = $n_0$Temp, $n_1^*$ = $n_1$Temp and $n_{01}^*$ = $n_{01}$Temp (= $n_0^*$ + $n_1^*$)
            Endif
      Endif
      % Continue the Inner While loop to examine the next x
   Endwhile (for Inner While loop)

   % Prepare for next iteration of the Outer While loop. The best solution found becomes
    the solution added to S.
   % If this point is reached because $R^o$ became empty in the Inner While loop, then the
outer loop automatically terminates.
      $z_j^*$ = 0 for j ∈ $N_0^*$; $z_j^*$ = 1 for j ∈ $N_1^*$ and $z_j^*$ = # otherwise

  % The preceding update for $z^*$, is implicit in $N_0^*$ and $N_1^*$.
  $x^* = x(h^*)$
  $S := S \cup \{x^*\}$
  $k := k + 1$ ($|S^*| = k$ for the current updated k)
  $N_0 = N_0^*$, $N_1 = N_1^*$ and $N_{01} = N_{01}^*$
  $n_0 = n_0^*$, $n_1 = n_1^*$ and $n_{01} = n_{01}^*$
  $n_v^* = 0$
  % This assures that the first non-empty set $N_v$ will qualify as a candidate to add to the new S.
Endwhile (for Outer While loop)

In summary, the method goes through all $x(h) \in R^o$ and picks the one that gives the largest $n_v$ value for $n_v = N_v(S \cup \{x\})$. This x is added to S if $n_v \geq L_v^1$, and the method terminates otherwise. (It is assumed that dropping an element from $R^o$ it does not affect the examination of remaining elements.) Each x is handled using temporary values of $N_0$, $N_1$ and $N_{01}$, since current values are needed for examining next x. The values of $N_0$, $N_1$ and $N_{01}$ only change when the best x is selected after examining all $R^o$.

The detailed algorithm for Problem $G_v^2$ employs a corresponding design, which can be inferred from the above and the description of the $G_v^2$ algorithm in Section 4.

## Appendix 2.

In the following elaboration of the algorithm illustrated in the example of Section 5, the vectors shown for the solutions $x(1)$ to $x(12)$ identify the z vectors generated for $v = 0$ when starting from $x^1 = x(1)$. After reaching $S = \{x(1), x(2), x(3)\}$ the z vectors for all remaining solutions $x(4)$ through $x(12)$ are shown as a basis for selecting one of the corresponding original $x(h)$ solutions to be added to S. As previously noted, the decrease in $|N_v(S)|$ from 5 to 3 (the maximum value available for expanding S by adding one more solution) suggests that 5 is a good cutoff value for $L_v^1$.

**Extended Table for v = 0:**

| | Vectors in R | | | | | | | | | | | | $|N_v(S)|$ for v = | | | |
|---|---|---|---|---|---|---|---|---|---|---|---|---|---|---|---|---|
| j = | 1 | 2 | 3 | 4 | 5 | 6 | 7 | 8 | 9 | 10 | 11 | 12 | 0 | 1 | 01 | S = |
| x(1) | 1 | 1 | 0 | 1 | 0 | 0 | 0 | 0 | 1 | 0 | 0 | 0 | **8** | 4 | 12 | {x(1)} |
| x(2) | # | # | # | # | # | 0 | 0 | 0 | # | 0 | 0 | 0 | **6** | 2 | 8 | {x(1), x(2)} |
| x(3) | # | # | # | # | # | 0 | # | 0 | # | 0 | 0 | 0 | **5** | 1 | 6 | {x(1), x(2), x(3)} |

(Evaluations for v = 0 for next iteration after the preceding)

| | | | | | | | | | | | | | | | | |
|---|---|---|---|---|---|---|---|---|---|---|---|---|---|---|---|---|
| x(4) | # | # | # | # | # | 1 | # | 1 | # | 0 | 0 | 0 | **3** | | | |
| x(5) | # | # | # | # | # | 1 | # | 0 | # | 0 | 1 | 0 | **3** | | | |

| | | | | | | | | | | | | |
|---|---|---|---|---|---|---|---|---|---|---|---|---|
| x(6) | # | # | # | # | # | 1 | # | 1 | # | 1 | 0 | 1 | **1** |
| x(7) | # | # | # | # | # | 1 | # | 1 | # | 0 | 1 | 1 | **1** |
| x(8) | # | # | # | # | # | 1 | # | 1 | # | 1 | 1 | 1 | **0** |
| x(9) | # | # | # | # | # | 0 | # | 1 | # | 1 | 1 | 1 | **1** |
| x(10) | # | # | # | # | # | 0 | # | 1 | # | 1 | 1 | 1 | **1** |
| x(11) | # | # | # | # | # | 0 | # | 1 | # | 1 | 0 | 1 | **2** |
| x(12) | # | # | # | # | # | 0 | # | 1 | # | 1 | 0 | 1 | **2** |

Next we identify the z vectors for the solutions x(1) to x(12) generated for v = 1 when starting from $x^1 = x(4)$. The vectors have been reordered for convenience so that x(4) appears first. After reaching S = {x(4), x(5), x(3)} the z vectors for all remaining solutions x(4) through x(12) are shown as a basis for selecting one of the corresponding original x(h) solutions to be added to S. Here, the decrease in $|N_v(S)|$ from 5 to 2 (the maximum value available for expanding S by adding one more solution) suggests that 5 is a good cutoff value for the size |S| of S, as represented by the lower bound $L_v^2$.

**Extended Table for v = 1:**

| | Vectors in R | | | | | | | | | | | | $|N_v(S)|$ for v = | | | S = |
|---|---|---|---|---|---|---|---|---|---|---|---|---|---|---|---|---|
| j = | 1 | 2 | 3 | 4 | 5 | 6 | 7 | 8 | 9 | 10 | 11 | 12 | 0 | 1 | 01 | |
| x(4) | 1 | 1 | 1 | 1 | 1 | 1 | 1 | 1 | # | # | # | # | 4 | **8** | 12 | {x(4)} |
| x(5) | 1 | 1 | 1 | # | 1 | 1 | 1 | # | # | # | # | # | 3 | **6** | 9 | {x(4), x(5)} |
| x(3) | 1 | 1 | 1 | # | 1 | # | 1 | # | # | # | # | # | 2 | **5** | 7 | {x(4), x(5), x(3)} |

(Evaluations for v = 1 for next iteration after the preceding)

| | | | | | | | | | | | | | |
|---|---|---|---|---|---|---|---|---|---|---|---|---|---|
| x(1) | 1 | 1 | 0 | # | 0 | # | 0 | # | # | # | # | # | **2** |
| x(2) | 0 | 0 | 1 | # | 1 | # | 0 | # | # | # | # | # | **2** |
| x(6) | 0 | 1 | 0 | # | 0 | # | 0 | # | # | # | # | # | **1** |
| x(7) | 0 | 1 | 0 | # | 0 | # | 0 | # | # | # | # | # | **1** |
| x(8) | 0 | 1 | 0 | # | 0 | # | 0 | # | # | # | # | # | **1** |
| x(9) | 0 | 0 | 0 | # | 1 | # | 1 | # | # | # | # | # | **2** |
| x(10) | 0 | 1 | 0 | # | 0 | # | 1 | # | # | # | # | # | **2** |
| x(11) | 1 | 0 | 0 | # | 0 | # | 1 | # | # | # | # | # | **2** |
| x(12) | 0 | 0 | 0 | # | 1 | # | 1 | # | # | # | # | # | **2** |